\documentclass[pdflatex,sn-mathphys-num]{sn-jnl}

\usepackage{graphicx}%
\usepackage{multirow}%
\usepackage{amsmath,amssymb,amsfonts}%
\usepackage{amsthm}%
\usepackage{mathrsfs}%
\usepackage[title]{appendix}%
\usepackage{xcolor}%
\usepackage{textcomp}%
\usepackage{manyfoot}%
\usepackage{booktabs}%
\usepackage{algorithm}%
\usepackage{algorithmicx}%
\usepackage{algpseudocode}%
\usepackage{listings}%
\usepackage{rotating}%
\usepackage{adjustbox}
\usepackage{tabularx}
\usepackage{booktabs}
\usepackage{makecell}
\usepackage{siunitx}
\usepackage{colortbl}
\usepackage[verbose]{placeins}
\usepackage{subcaption}
\usepackage[capitalize]{cleveref}
\crefname{section}{Sec.}{Secs.}
\Crefname{section}{Sec.}{Sections}
\Crefname{table}{Table}{Tables}
\crefname{table}{Table}{Tabs.}
\usepackage{float}
\usepackage{bbold}

\newcommand{\figref}[1]{Fig.~\ref{#1}}
\newcommand{\tabref}[1]{Tab.~\ref{#1}}
\newcommand{\eqnref}[1]{Eqn.~\ref{#1}}
\newcommand{\secref}[1]{Sec.~\ref{#1}}
\newcommand{\apdxref}[1]{Appendix~\ref{#1}}

\begin{document}

\title[Article Title]{Farm-Level, In-Season Crop Identification for India}


\author*[1] {\fnm{Ishan} \sur{Deshpande}}\email{ishansd@google.com}
\author[3] {\fnm{Amandeep Kaur} \sur{Reehal}}
\author[4] {\fnm{Chandan} \sur{Nath}}
\author[4] {\fnm{Renu} \sur{Singh}}
\author[2] {\fnm{Aayush} \sur{Patel}}
\author[1] {\fnm{Aishwarya} \sur{Jayagopal}}
\author[4] {\fnm{Gaurav} \sur{Singh}}
\author[4] {\fnm{Gaurav} \sur{Aggarwal}}
\author[2] {\fnm{Amit} \sur{Agarwal}}
\author[1] {\fnm{Prathmesh} \sur{Bele}}
\author[2] {\fnm{Sridhar} \sur{Reddy}}
\author[4] {\fnm{Tanya} \sur{Warrier}}
\author[2] {\fnm{Kinjal} \sur{Singh}}
\author[2] {\fnm{Ashish} \sur{Tendulkar}}
\author[2] {\fnm{Luis Pazos} \sur{Outon}}
\author[1] {\fnm{Nikita} \sur{Saxena}}
\author[1] {\fnm{Agata} \sur{Dondzik}}
\author[1] {\fnm{Dinesh} \sur{Tewari}}
\author[1] {\fnm{Shruti} \sur{Garg}}
\author[1] {\fnm{Avneet} \sur{Singh}}
\author[1] {\fnm{Harsh} \sur{Dhand}}
\author[1] {\fnm{Vaibhav} \sur{Rajan}}
\author[1] {\fnm{Alok} \sur{Talekar}}

\affil[1]{\orgdiv{Google DeepMind}}
\affil[2]{\orgdiv{Google}}
\affil[3]{\orgdiv{Work done while at Google DeepMind}}
\affil[4]{\orgdiv{Work done while at Google}}




\abstract{Accurate, timely, and farm-level crop type information is paramount for national food security, agricultural policy formulation, and economic planning, particularly in agriculturally significant nations like India. While remote sensing and machine learning have become vital tools for crop monitoring, existing approaches often grapple with challenges such as limited geographical scalability, restricted crop type coverage, the complexities of mixed-pixel and heterogeneous landscapes, and crucially, the robust in-season identification essential for proactive decision-making. A primary impediment has been the scarcity of extensive, consistent ground-truth data, especially regarding precise crop phenology (sowing and harvest dates), which limits the development and validation of truly operational in-season systems at a national scale.
\\
\\
We present a framework designed to address the critical data gaps for targeted data driven decision making which generates farm-level, in-season, multi-crop identification at national scale (India) using deep learning. Our methodology leverages the strengths of Sentinel-1 and Sentinel-2 satellite imagery, integrated with national-scale farm boundary data. The model successfully identifies 12 major crops (which collectively account for nearly 90\% of India's total cultivated area showing an agreement with national crop census 2023-24 of 94\% in winter, and 75\% in monsoon season). Our approach incorporates an automated season detection algorithm, which estimates crop sowing and harvest periods. This allows for reliable crop identification as early as two months into the growing season and facilitates rigorous in-season performance evaluation. Furthermore, we have engineered a highly scalable inference pipeline, culminating in what is, to our knowledge, the first pan-India, in-season, farm-level crop type data product. The system's effectiveness and scalability are demonstrated through robust validation against national agricultural statistics, showcasing its potential to deliver actionable, data-driven insights for transformative agricultural monitoring and management across India.}
\maketitle

\section{Introduction}\label{sec:intro}
\noindent The agricultural sector forms the bedrock of global food security and economic stability, with its significance magnified in nations like India, where it underpins the livelihoods of a substantial portion of the population and contributes significantly to the national economy~\cite{chand2022indian, faoindiaataglance}. The sheer scale and complexity of India's agricultural landscape—characterized by diverse agro-climatic zones, a predominance of smallholder farms, varied cropping patterns, and intricate traditional practices—present formidable challenges to effective monitoring and management~\cite{faoindiaataglance}. Accurate and timely information on crop types, particularly at the farm level and within the growing season, is indispensable for informed policy-making, efficient resource allocation, proactive food security measures, and the sustainable intensification of agriculture~\cite{gao2023training}. In-season insights, specifically, enable forward-looking estimates of crop production, facilitate timely interventions in response to climatic stressors or pest outbreaks, and support the logistical planning of numerous agro-support industries, including insurance, input supply chains, and storage~\cite{zhao2022season, gao2023training}.
\\
\\
\noindent For several decades, remote sensing technology, coupled with advancements in machine learning, has been an active area of research for crop identification~\cite{soler2024combining, yao2022classification, zhang2022towards, xie2021crop, you2020examining}. The fundamental problem involves leveraging satellite observations to determine the specific crop cultivated within a given farm boundary, either post-season or, more critically, in-season. Traditional approaches often treat this as a classification task, training models to distinguish between various crop categories based on their spectral and temporal signatures derived from satellite imagery~\cite{kerner2020resilientinseasoncroptype, lin2022early, Wang2023CropformerAN, weilandt2023early, tseng2024lightweightpretrainedtransformersremote, agronomy13071723, chang2024generalizability}. However, many existing solutions face significant limitations when considered for operational, large-scale deployment in complex environments like India. These limitations often include a restricted focus on a few dominant crop types~\cite{turkoglu2021crop}, applicability confined to small geographical regions due to dependencies on localized training data~\cite{turkoglu2021crop, reuss2025eurocropsml, russwurm2019breizhcrops}, and a lack of robust methodologies for true in-season identification and validation across different phenological stages. A persistent bottleneck has been the availability of extensive, high-quality, and temporally precise ground-truth data, especially concerning actual sowing and harvest dates, which is crucial for training and evaluating models capable of reliable in-season performance~\cite{soler2024combining}.
\\
\\
\noindent This work aims to bridge these data gaps by developing a farm-level, in-season multi-crop identification data product that is available at national scale, and at a high temporal frequency to enable timely decision making. Building upon a foundational understanding of the agricultural landscape, which includes the delineation of individual farm boundaries, our approach integrates multi-temporal data from Sentinel-1 and Sentinel-2 satellites within a deep learning framework. We successfully identify 12 major crops that collectively cover nearly 90\% of India's cultivated area. Our method explicitly incorporates a season detection algorithm, which estimates crop sowing and harvest dates at the field level. This enables our model to reliably identify crops as early as two months into their growing season and allows for a more nuanced evaluation of its performance as the season progresses. Furthermore, we have developed a highly scalable inference pipeline designed to generate these farm-level predictions for all of India, representing, to our knowledge, the first such comprehensive in-season crop type data product at this scale for the nation. The efficacy of our system is verified through rigorous comparisons of aggregated predictions against official agricultural statistics, demonstrating its potential to provide actionable, data-driven insights for transformative agricultural monitoring and management.
\\
\\
\noindent To summarize, our contributions are as follows:
\begin{itemize}
\item We provide a first-of-its-kind pan-India, field-level, multi-crop dataset, which is updated at a high temporal cadence.
\item We utilise season information in predicting in-season crop labels, by incorporating a season detection algorithm in our pipeline.
\item We perform rigorous evaluation of our model's predictions against national crop census data (2023-24).
\end{itemize}

\noindent In the remainder of this paper, we discuss our method and contributions in more details. The paper is organized as follows. We first discuss related works in \secref{sec:background}. We then present a discussion on our model, the ground truth data, and training methodology in \secref{sec:method}. This is followed by evaluation of the model (farm-level and aggregate). We then conclude with a brief discussion of future directions.

\section{Background}
\label{sec:background}
\textbf{Preliminaries:} A \textit{crop season} refers to the time period during which a crop is active on a farm, i.e. the time from sowing till harvesting. \textit{In-season} crop classification refers to classifying the crop type for an active crop season, meaning that when making predictions for a certain date within the crop season no information after that date must be used as input. This is in contrast with \textit{post-season} crop classification where there is no such time constraint on the inputs used. While post-season classification is useful in its own right for after-the-fact analysis historical tracking, in-season classification is necessary for near-term forecasting and associated decisions. \\
\\
\textbf{Remote Sensing Base Solutions:} Because of the nature of the problem, in particular the geographical scale at which any solution to it must be applied, remote sensing data (i.e. satellite observations) is particularly convenient to use as input. Furthermore, several satellites capture multi-spectral information which is useful for differentiating between crops. We now have several Earth-observation satellites continuously monitoring the Earth and providing us with a stream of information. Methods based on remote sensing can be easily applied to vast areas, as opposed to methods such as manual surveys which are expensive and time-consuming. This has spurred extensive research into the use of satellite based methods for crop identification, particularly as the data has become publicly accessible through platforms like Earth Engine (\cite{gorelick2017google}). The LANDSAT satellites were popular inputs for early methods (\cite{CAI201835, landsatmodis, landsatperu, 8899274}). For most of the recent methods, Sentinel-2 (\cite{spoto2012overview}) and Sentinel-1 (\cite{torres2012gmes}) are the main sources of information and are often augmented with LANDSAT observations (\cite{BLICKENSDORFER2022112831, kerner2020resilientinseasoncroptype, lin2022early, Wang2023CropformerAN, weilandt2023early, tseng2024lightweightpretrainedtransformersremote, agronomy13071723, chang2024generalizability, Orynbaikyzy02092019}). We also follow this trend and use the Sentinel satellites in our approach. See \secref{sec:satellite_data} for a longer discussion on satellite observations. \\
\\
\textbf{Datasets:} There are several datasets with farm-level or lat/lng-level crop maps, such as \cite{turkoglu2021crop, russwurm2019breizhcrops, reuss2025eurocropsml, tseng2021cropharvest, kerner2025fields, Boryan01082011}. These typically provide information including farm boundary/ coordinates, crop label, and year. Most prior works use one of these datasets, or a proprietary dataset, and couple it with satellite observations for training models. However, these datasets are not suitable for our use case for two reasons. Firstly, these datasets are taken either from the US or Europe. We instead focus on India, where the major crops are different and the crop signatures might also differ. Secondly, these datasets do not contain corresponding sowing and harvest dates, and instead only provide the crop label for the crop that was growing for the major part of the year. Therefore, the information in these datasets is not sufficient by itself to train models that can perform in-season classification. Therefore, in our work we license and use a dataset that was collected in India. We use season detection (\cite{season_detection}) to estimate sowing and harvest dates that allows us to train for in-season classification.\\
\\
\textbf{Modeling Methods:} Typically, the satellite observations are used to construct a time-series to capture the temporal dynamics of crops. The observations either cover a single location (\cite{tseng2021learning, chang2024generalizability, gurav2023can, yao2022classification, zhang2022towards, xie2021crop, lin2022early, you2020examining}), the entire farm (\cite{mahlayeye2022cropping, object_based1, object_based2, weilandt2023early, abdali2023parallel, yin2018mapping}), or a large region including and surrounding the farm (\cite{kerner2020resilientinseasoncroptype, lin2022early, Wang2023CropformerAN, weilandt2023early, tseng2024lightweightpretrainedtransformersremote, agronomy13071723, chang2024generalizability}). It has been observed that using information that covers the entire farm is typically better than using information from just a single location. Following this observation, we use farm-level information in our approach. Popular choices of machine learning models include random forests (\cite{soler2024combining, yao2022classification, zhang2022towards, xie2021crop, you2020examining}), CNNs (\cite{lin2022early, kerner2020resilientinseasoncroptype, agronomy13071723, chang2024generalizability}), and transformers (\cite{tseng2024lightweightpretrainedtransformersremote, Wang2023CropformerAN, weilandt2023early}). Several recent works like \cite{tseng2024lightweightpretrainedtransformersremote, Wang2023CropformerAN, 9252123, chang2024generalizability} leverage unsupervised learning as a means to pre-train large transformer models, as this approach has shown promise in several other domains. We follow this larger trend towards a combination of unsupervised and supervised learning with transformer based models.\\
\\
\textbf{Farm boundaries:} As mentioned earlier, we use farm-level information as input for our method. For this we use the farm boundaries predicted by the Agricultural Landscape Understanding model \textbf{ALU} (\cite{dua2024agriculturallandscapeunderstandingcountryscale}), which infers the farm boundaries at 1m resolution by using a deep learning model with high res satellite imagery.

\section{Method}
\label{sec:method}
In this section, we discuss the complete pipeline we use for crop classification using satellite data. We use a two stage approach - the first stage identifies active crop seasons, and the second stage classifies the crop for active crop seasons. This separation of stages allows us to decouple and simplify problems. For the first stage, we use the season detection algorithm from \cite{season_detection} without modifications. It is discussed in more detail in \apdxref{appendix:season_detection}. In the rest of this paper we focus on the second stage - crop classification \textit{for active crop seasons}. Each component of the crop classification pipeline, including the data, the pre-processing, and the model architecture, are discussed separately.

\subsection{Ground Truth Labels}
We use a dataset of crop labels licensed from multiple partners for training and evaluating our models where each sample in the dataset contains the following information:
\begin{itemize}
    \item \textbf{Crop}: The crop label.
    \item \textbf{Coordinates}: The location for the label, as a (latitude, longitude) tuple.
    \item \textbf{Timestamp}: The timestamp of label collection.
\end{itemize}

\begin{table}[]
\caption{Number of GT Samples and Unique S2 Cells Covered}
\label{tab:crop_label_counts_histogram}
\begin{tabular}{@{}lcc@{}}
\toprule
\textbf{Crop} & \textbf{Count} & \textbf{Num S2 Cells @ Level-9} \\ \midrule
\multicolumn{1}{|l|}{Wheat} & \multicolumn{1}{c|}{14,072} & \multicolumn{1}{c|}{995} \\ \midrule
\multicolumn{1}{|l|}{Sugarcane} & \multicolumn{1}{c|}{9,436} & \multicolumn{1}{c|}{233} \\ \midrule
\multicolumn{1}{|l|}{Soybeans} & \multicolumn{1}{c|}{7,001} & \multicolumn{1}{c|}{641} \\ \midrule
\multicolumn{1}{|l|}{Mustard} & \multicolumn{1}{c|}{6,598} & \multicolumn{1}{c|}{477} \\ \midrule
\multicolumn{1}{|l|}{Corn} & \multicolumn{1}{c|}{5,499} & \multicolumn{1}{c|}{537} \\ \midrule
\multicolumn{1}{|l|}{Rice} & \multicolumn{1}{c|}{4,888} & \multicolumn{1}{c|}{620} \\ \midrule
\multicolumn{1}{|l|}{Cotton} & \multicolumn{1}{c|}{4,365} & \multicolumn{1}{c|}{465} \\ \midrule
\multicolumn{1}{|l|}{Gram (Bengalgram)} & \multicolumn{1}{c|}{4,170} & \multicolumn{1}{c|}{443} \\ \midrule
\multicolumn{1}{|l|}{Sorghum} & \multicolumn{1}{c|}{3,635} & \multicolumn{1}{c|}{323} \\ \midrule
\multicolumn{1}{|l|}{Groundnut} & \multicolumn{1}{c|}{2,509} & \multicolumn{1}{c|}{163} \\ \midrule
\multicolumn{1}{|l|}{Chilli} & \multicolumn{1}{c|}{1,993} & \multicolumn{1}{c|}{112} \\ \midrule
\multicolumn{1}{|l|}{Bajra (Pearl Millet)} & \multicolumn{1}{c|}{1,732} & \multicolumn{1}{c|}{176} \\ \midrule
\multicolumn{1}{|l|}{Others (63 crops)} & \multicolumn{1}{c|}{3,825} & \multicolumn{1}{c|}{859} \\ \midrule
\textbf{Total} & \textbf{69,723} & \textbf{2,097} \\ \bottomrule
\end{tabular}
\end{table}

\noindent The dataset contains a total of $\sim$ 70,000 samples.
The labels in the samples include a total of 75 crops, covering most of the major crops grown in India. Distribution of the labels and the number of S2 cells spanned per label is shown in \tabref{tab:crop_label_counts_histogram}. Notably, we have a very long-tailed distribution with 12 crops accounting for $\sim 95\%$ of the dataset and the remaining 63 accounting for $\sim 5\%$ .\\

\noindent Different aspects of the dataset, including the label sourcing processing, geographic distribution, and temporal coverage are discussed in \apdxref{appendix:ground_truth_data}.

\subsection{Training Data Generation}
\label{sec:training_data_generation}
In this section we discuss how we generate training inputs for the model corresponding to the ground truth labels. Our model uses Sentinel-1 and Sentinel-2 signals as inputs for prediction. We first briefly describe the satellite signals and then discuss the training data generation.

\subsubsection{Satellite Data}
\label{sec:satellite_data}
Sentinel-1 and Sentinel-2 provide periodic streams of observations covering the entire surface of the Earth, with a time period varying between 3 to 8 days\footnote{The exact frequency varies depends on the latitude, with a higher frequency near the poles and lower frequency near the equator.} between two observations. The observations are captured at various spatial resolutions, ranging from 10m to 60m. We re-sample all observations to a resolution of 10m, meaning that if we impose a 2-d grid of points on the Earth's surface such that the inter-point spacing along either axis is 10m, then information is assumed to be independently available for each point on this grid (just like any image-like signal).\\
\\
Therefore, for any point on the aforementioned grid, we have a time-series of observations. An observation itself consists of information collected across multiple ``bands'' (similar to the typical RGB, but more generalized), where the information in a single band is represented with a single float. For each point in the grid, for each satellite $s$ we construct a feature vector $f^s_t$ for time-step $t$ by concatenating the value of all bands at time $t$ together, i.e.  $f^s_t = (b^1_t, b^2_t,...,b^N_t)$ where $b^i_t$ is the value of the $i^{th}$ band of satellite $s$ at time $t$.\\
\\
Therefore, for any location on the Earth's surface we will have multiple independent time-series of features, one from each satellite. We collectively refer to the satellite information at a location $(x, y)$ on the Earth's surface between time $t_{start}$ and $t_{end}$ as $\text{\textbf{RSD}}_{(t_{start}, t_{end})}^{x,y}$ (Remote Sensing Data). If we have $M$ satellites $\text{S}_1$, $\text{S}_2$, ..., $\text{S}_M$, then:
\begin{equation}
\label{eq:rsd}
    \text{\textbf{RSD}}_{(t_{start}, t_{end})}^{x,y} := \{(f^{\text{S}_i}_{t_1}, f^{\text{S}_i}_{t_2}, ..., f^{\text{S}_i}_{t_N})\}_{i = 1}^M,
\end{equation}
where $t_{start} = t_1 < t_2 < ... < t_N = t_{end}$. In our case $M = 2$ as we use two satellites, but this can be extended to use any arbitrary collection of satellites. Note that we will drop the subscript from the notation when not specifying a finite time-period.

\subsubsection{Generation Process}
\label{sec:generation_process}
\begin{figure}[t]
    \centering
    \includegraphics[width=0.9\columnwidth]{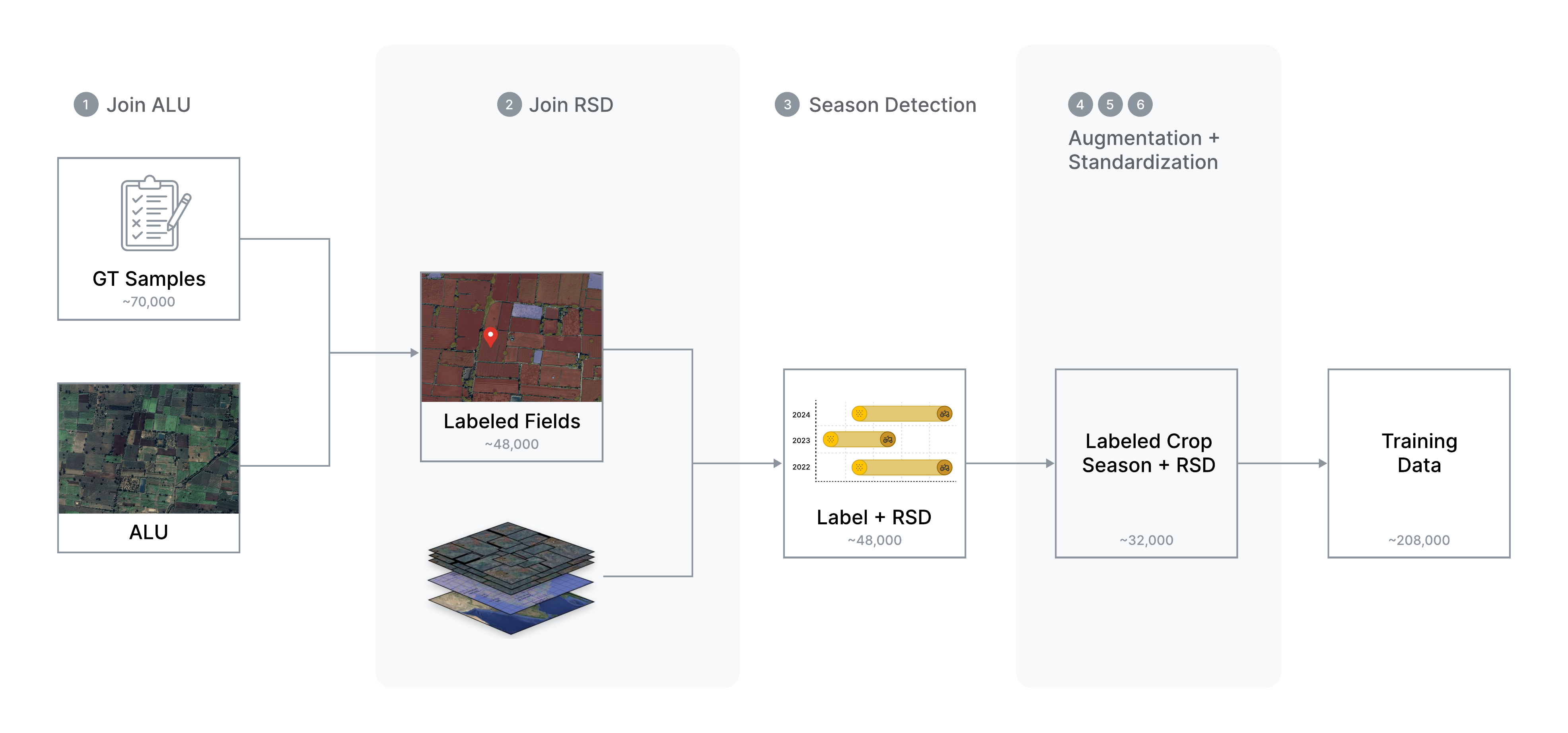}
    \caption{\textbf{Training Data Generation:} \small{Our training data generation process combines the labels with satellite data from Earth Engine and field boundaries from ALU. This is followed by season detection to estimate sowing and harvesting dates. Finally, the data goes through normalization and augmentation.}}
    \label{fig:train_gen}
\end{figure}

The process of generating the inputs, visualized in \figref{fig:train_gen}, consists of the following sequential steps: 
\begin{itemize}
    \item[1.] \textbf{Join with ALU:} We use the field boundaries from ALU (\cite{dua2024agriculturallandscapeunderstandingcountryscale}) and associate each sample with a field by checking if the coordinates of the sample lie inside the ALU boundary. One field can match with multiple samples. We ignore samples that do not match with any fields from ALU. In this stage several samples are dropped because either the collected sample had a location outside the field, or the ALU boundaries are incorrect.\\ 
    Output: \textit{labeled fields} $=$ (field, labels).\\
    Num Samples Retained: 48,285.
    \item[2.] \textbf{Join with RSD:} We join each labeled field with RSD (\eqnref{eq:rsd}) for all points in the field sampled at a 10m resolution which lie at least 10m inside the field boundary. The RSD is obtained from Earth Engine (\cite{gorelick2017google}). The check on the distance from the boundary is important for small fields because otherwise the RSD might contain information from outside the field.  This gives us a collection of RSDs for each field, ${RSD^{x_i,y_i}}$ where ${(x_i, y_i)}$ are the interior points in the field. Then we construct RSD for the field by firstly interpolating the RSD of each point in time to get a uniform temporal frequency of 5 days, and then taking the median of features for each timestep over all points in the field as suggested in \cite{object_based1} and \cite{object_based2}. We call this as $\text{\textbf{RSD}}^{field}$. \\
    Output: \textit{labels with satellite data} $=$ (labels, $\text{\textbf{RSD}}^{field}$).\\
    Num Samples Retained: 48,285
    \item[3.] \textbf{Season Detection:} We use the season detection algorithm from \cite{season_detection} to identify the crop season bounds associated with each sample, defined as \textbf{(estimated season start time, estimated season end time)}. We discard all samples that do not have a uniquely identified season (either no detected season covering the label or overlapping seasons with different labels). We also discard any samples where the detected season length does not match normative season lengths for the crop label (details in \apdxref{appendix:season_length}). A large fraction of the remaining samples are dropped in this stage. This can be attributed to either the ground truth being collected outside of the actual crop season, or the season detection algorithm not being able to identify the season correctly.\\
    Output: \textit{labeled crop season} $=$ (season start time, season end time, label, $\text{\textbf{RSD}}^{field}$).\\
    Num Samples Retained: 32,208\\
    \textit{Note: More than half of the labeled data is discarded cumulatively across step 1 and 3 due to data and ALU quality.}
    \item[4.] \textbf{Temporal Augmentation:} Since by definition of the crop season the same crop label applies throughout the season, we generate multiple samples per ground truth sample by creating ``augmented'' labels with varying label timestamps within the season. Concretely, we sample $t_{end}$ at a 30 day interval from \textbf{(season start time, season end time)}. Then for each $t_{end}$ we generate a new sample which we refer to as an augmented sample. In this paper we refer to the difference between $t_{end}$ and the estimated season start as \textbf{days after estimated season start}. \\
    Output: \textit{augmented samples} $=$ (label, $t_{end}$, $\text{\textbf{RSD}}^{field}$)\\
    Num Augmented Samples: 208,143
    \item[5.] \textbf{In-Season Example:} We temporally slice the RSD for each sample so that (1) there is no information available from after $t_{label}$, and (2) exactly 1 year of information is available in RSD. In other words, we take
    \begin{equation}
        \text{\textbf{RSD}}^{in-season} = \text{\textbf{RSD}}^{field}_{(t_{end} - \text{1 year}, t_{end})}.
    \end{equation}
    This ensures that all examples have the labeled crop at the \textit{end} of the time series, making it easy to train the model to focus only on the active crop season. Additionally, we believe that the 1 year time period provides enough information about the current active crop season.\\ 
    Output: \textit{in-season examples} $=$ (label, $\text{\textbf{RSD}}^{in-season}$)\\
    Num In-Season Examples: 208,143
    \item[6.] \textbf{Standardization:} Finally, we apply the usual standardization for each in-season example, including z-score normalization and padding.\\
    Output: \textit{in-season examples} $=$ (label, $\text{\textbf{RSD}}^{in-season}$)\\
    Num In-Season Examples: 208,143
\end{itemize}

\noindent An important aspect of the training data is that we do not include any location or absolute time related information as inputs to the model, as this may induce unwanted priors in the model. Instead, we want the model to focus only on the satellite signal which will allow the model to generalize better. Any region/ time based priors can be included later during post-processing of outputs.

\subsubsection{Dataset Splits}
\label{sec:dataset_splits}
We split the dataset into training, validation, and test splits by sharding the data by location. Specifically, we group the data by level-9 S2 cells\footnote{A level-9 S2 cell, on average, covers $\sim 300$ sq. km area. See \cite{S2CellStatistics} for details.} and assign each S2 cell randomly to one of the splits, following a ratio of 70:15:15. This ensures that there is no ``leakage" of information across splits, especially through noise patterns introduced by weather which may similarly affect multiple nearby locations (e.g. clouds).

\subsection{Model Architecture}
\begin{figure}[h]
    \centering
    \includegraphics[width=0.99\columnwidth]{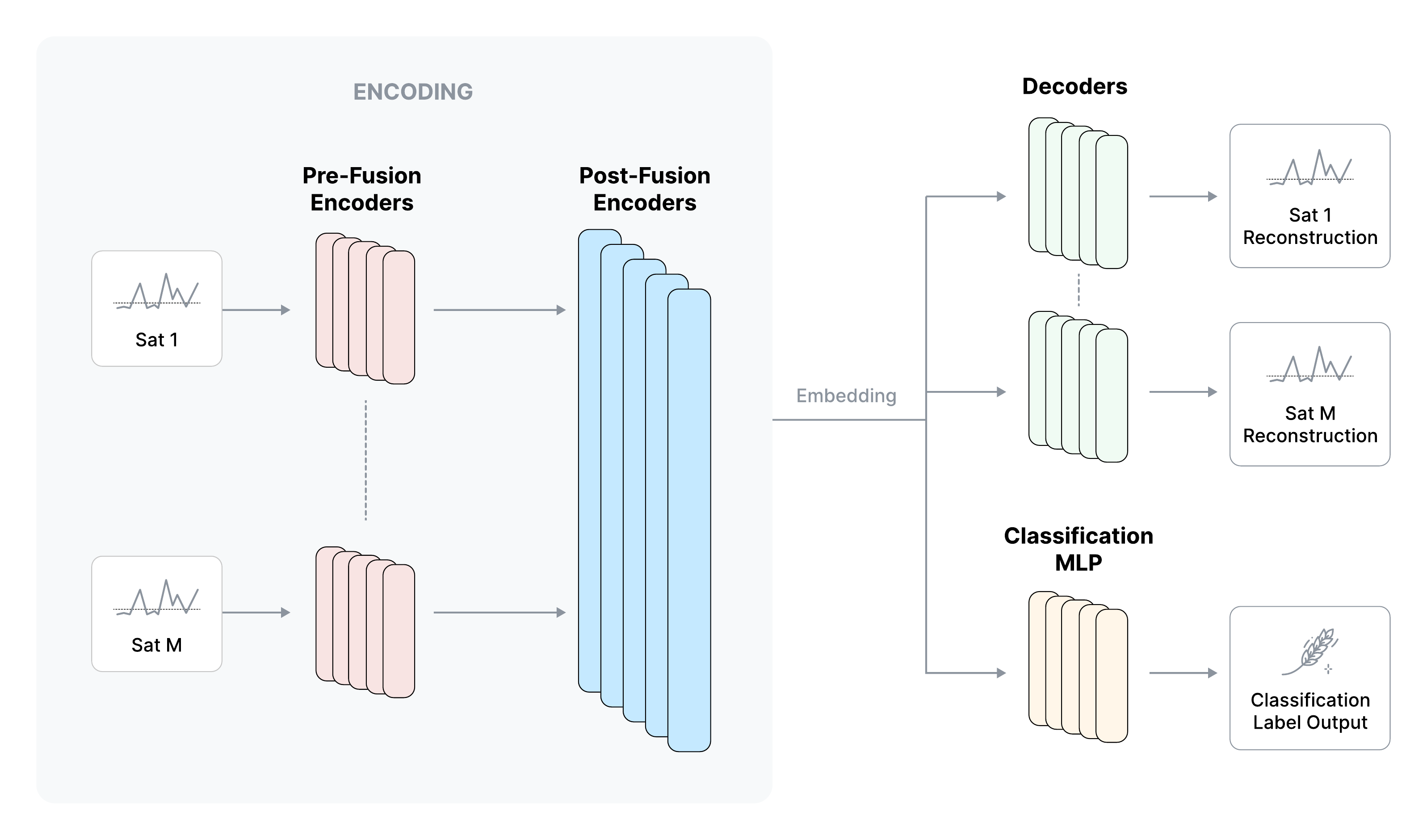}
    \caption{\textbf{Model Architecture:} \small{Our model consists of a transformer based encoder with dedicated attention layers per satellite as well as shared attention layers. We use a broadcast decoder for pre-training with MAE, and a separate MLP head for crop classification.}}
    \label{fig:model_architecture}
\end{figure}

We use a transformer based model (\cite{vaswani2023attentionneed}, \cite{xu2023multimodallearningtransformerssurvey}) for classification. The model, visualized in \figref{fig:model_architecture}, has an encoder-decoder architecture along with an MLP classification head. The encoder consists of independent tokenizers and self-attention layers for each satellite (referred to as pre-fusion encoders), followed by common attention layers (referred to as post-fusion encoders) that attend to all satellites together. The encoders use standard index based position encodings as we do not provide actual time related information. A temporal broadcast decoder based on \cite{watters2019spatialbroadcastdecodersimple} is used per satellite that decodes each time step of each satellite independently. The model is trained in two stages:
\begin{itemize}
\item \textbf{Pre-training:} The encoders and decoders are trained on unlabeled data using the MAE approach. We randomly mask out a part of the signals and train the model for reconstruction like \cite{li2023timaeselfsupervisedmaskedtime} and \cite{he2021maskedautoencodersscalablevision}. The shared encoder layers, masking, and reconstruction together encourage cross satellite learning. Details on the pre-training, including data, checkpoint selection, and hyper-parameters are discussed in \apdxref{appendix:model_training}.
\item \textbf{Fine-tuning:} The (pre-trained) encoder and the (un-trained) classification head are then trained for direct classification on the labeled training data. Since the training data generation process ensures that the labeled crop is at the end of the time-series, the model automatically learns to predict the active crop given the inputs. See \apdxref{appendix:model_training} for additional details.
\end{itemize}

\section{Evaluation}
\label{sec:evaluation}
\begin{figure}[t]
    \centering
    \includegraphics[width=\linewidth]{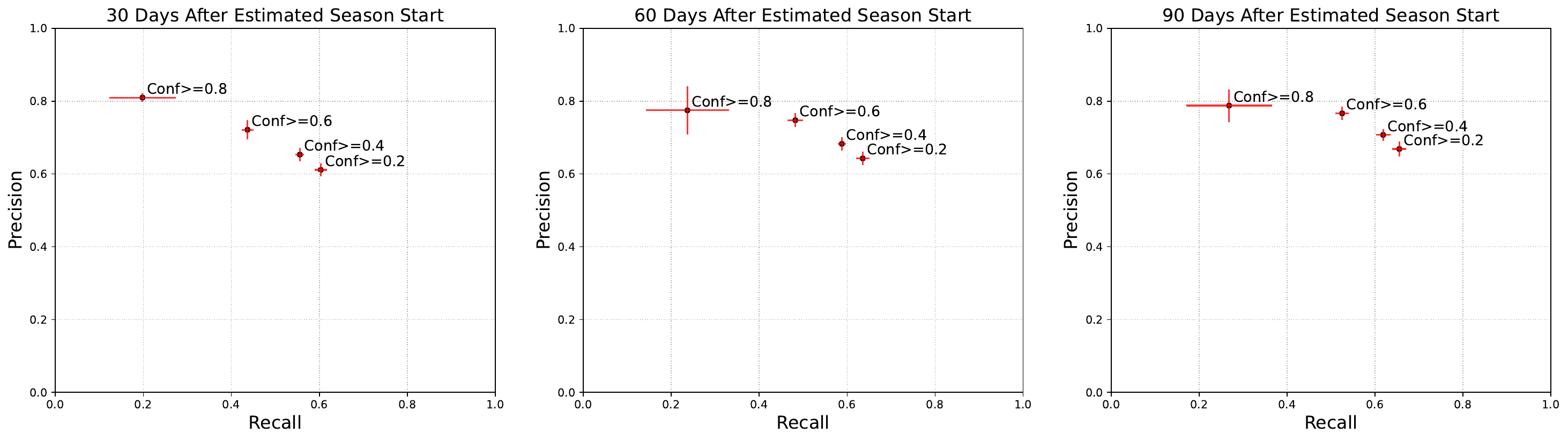}
    \caption{Precision, Recall vs Confidence Through the Crop Season}
    \label{fig:precision_recall_curve}
\end{figure}

\begin{figure}[H]
\centering

\begin{subfigure}{.95\textwidth}
  \caption{Winter (Rabi)}
  \label{fig:precision_winter}
  \centering
  \includegraphics[width=\linewidth]{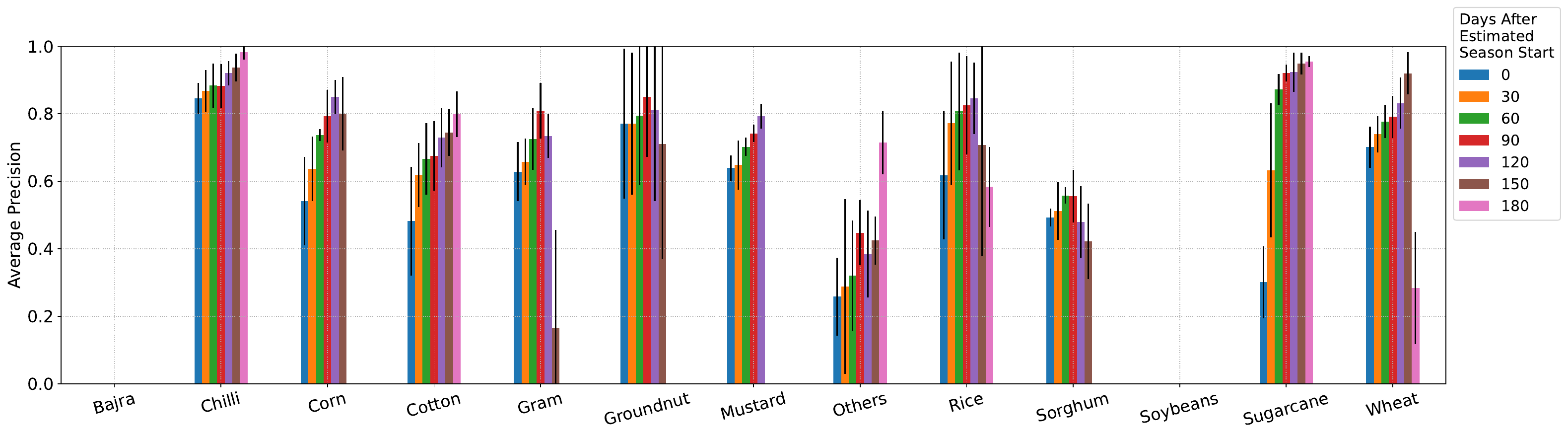}
\end{subfigure}

\begin{subfigure}{.95\textwidth}
  \caption{Monsoon (Kharif)}
  \label{fig:precision_monsoon}
  \centering
  \includegraphics[width=\linewidth]{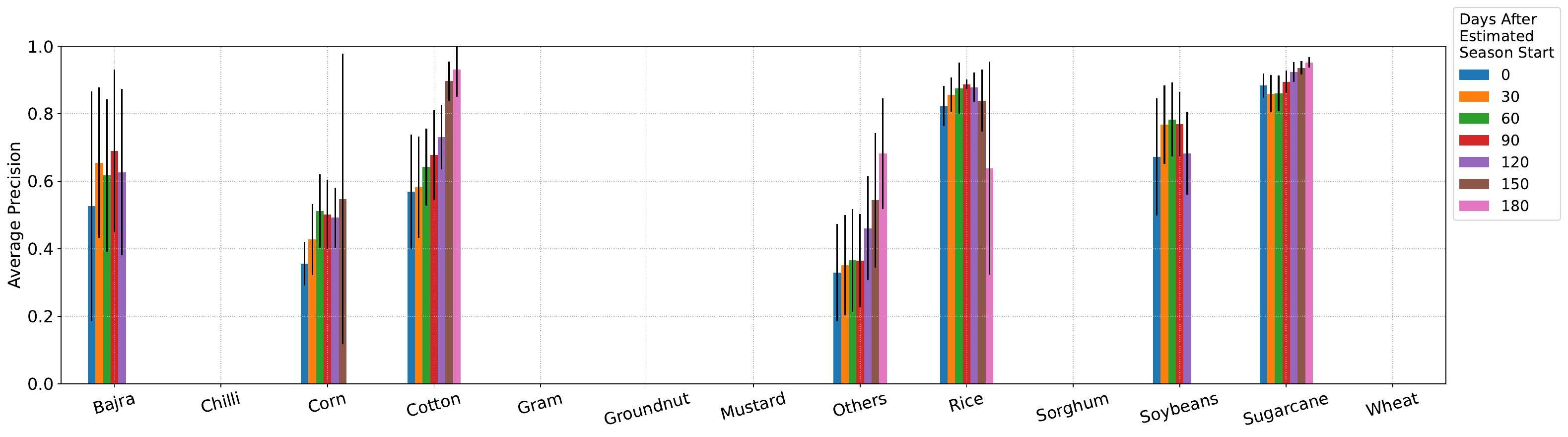}
\end{subfigure}%

\caption{\textbf{Per-crop Precision over Time:} \small{The figures above shows the precision of our model, averaged over three random splits, as it evolves through the crop season. Performance is measured separately for each season.}}
\label{fig:precision}
\end{figure}

\begin{figure}[H]
\centering
\begin{subfigure}{.95\textwidth}
  \centering
  \caption{Winter (Rabi)}
  \label{fig:recall_winter}
  \includegraphics[width=\linewidth]{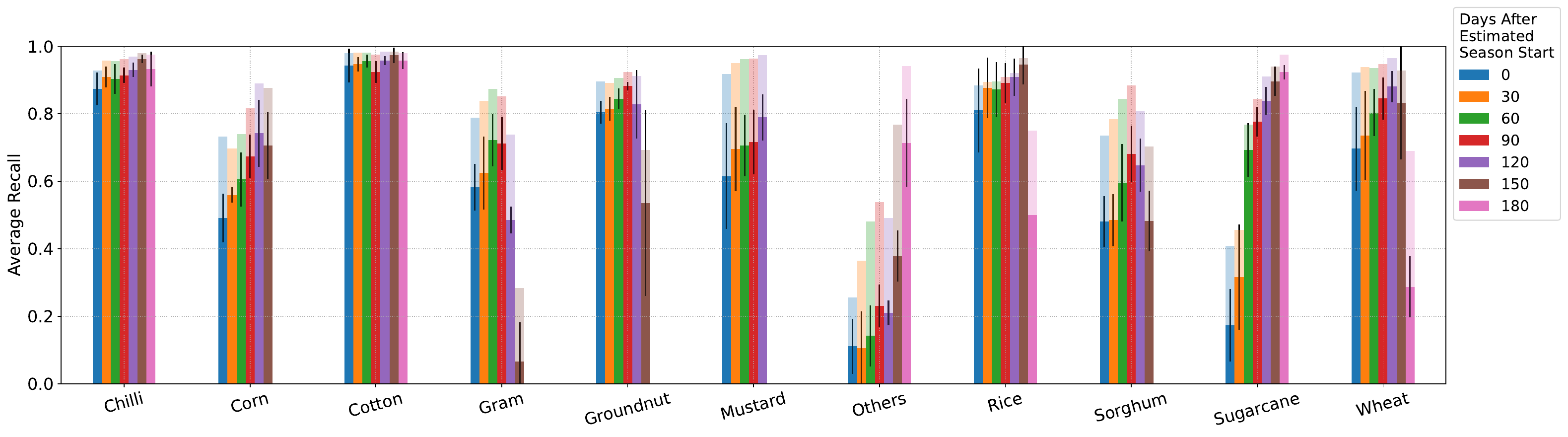}
\end{subfigure}

\begin{subfigure}{.95\textwidth}
    \caption{Monsoon (Kharif)}
  \label{fig:recall_monsoon}
  \centering
  \includegraphics[width=\linewidth]{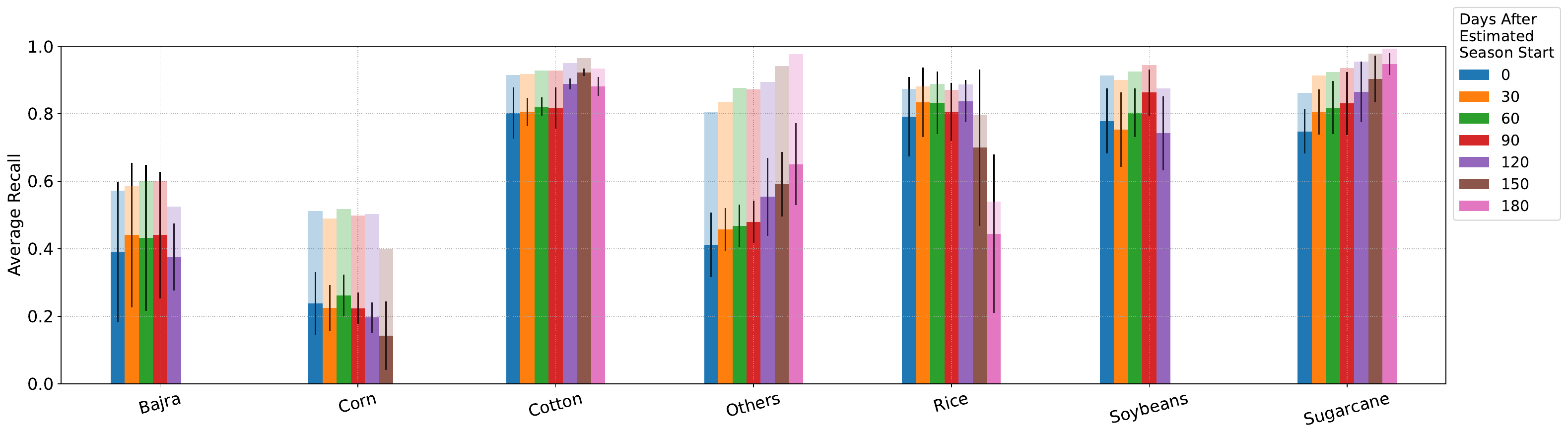}
\end{subfigure}

\caption{\textbf{Per-crop Recall over Time:} \small{The figures above shows the recall of our model, averaged over three random splits, as it evolves through the crop season. Performance is measured separately for each season. The solid bars show the recall in the most confident prediction. The translucent bars show the recall within the top-2 most confident predictions.}}
\label{fig:recall}
\end{figure}

In this section, we present evaluation results for the trained models. Our evaluations cover two separate dimensions that are indicative of the important use cases of in-season crop classification models:
\begin{itemize}
    \item[1.] Field-level predictions, and
    \item[2.] Aggregate predictions across large geographical areas. 
\end{itemize}

\noindent We first discuss details of how we train and select the models for evaluation, followed by a discussion of field level evaluation with labeled data, and finally an aggregated evaluation against census data.

\subsection{Model Training \& Selection}
\label{sec:evaluation_model_training_selection}
\noindent\textbf{Labels:} We keep the top 12 most frequent labels as is, and group all other labels into a single ``OTHERS'' label.\\
\\
\noindent\textbf{Training:} We generate three versions of the dataset splits with different random seeds, following the procedure from \secref{sec:dataset_splits}. We train models separately for each version of the dataset using the respective training split. The best model per version is selected with the relevant validation split. Specifically, we pick the model with the highest macro-average F1 score on the validation set per version, resulting in three separate models.\\
\\
\noindent\textbf{Evaluation:} Each of the three final models is evaluated on the appropriate test split for computing the final metrics.

\subsection{Field-level Evaluation with Labeled Data}
\label{sec:evaluation_with_labels}
\noindent We present evaluation results on the tests splits for the three models selected in \ref{sec:evaluation_model_training_selection}. Since we are interested in the in-season performance of the models, we assess the prediction accuracy at different points in the crop season, as measured by the difference in the prediction time and the start of the crop season. This is referred to as ``Days after Estimated Season Start'' in our evaluation. Note that this is only an \textit{estimate} as we are relying on the season detection algorithm for getting the season start/end dates.\\
\\
We use per-crop precision and recall as the primary metrics. The choice of metrics is driven by the fact that our dataset is highly imbalanced (thereby making raw accuracy score not viable) and our downstream use cases, where users might be interested in performance w.r.t particular crops as opposed to aggregate metrics. Precision enables users to estimate what fraction of fields predicted to have a certain crop can be expected to have that crop, allowing verification tasks. Per-crop recall helps us understand what fraction of fields that actually grow a certain crop can be identified by our model. Mathematically, these metrics are defined in terms of true positives $TP$, false positives $FP$, and false negatives $FN$, as:
\begin{equation}
    \text{Precision} = \frac{TP}{TP + FP},
\end{equation}
and 
\begin{equation}
    \text{Recall} = \frac{TP}{TP + FN}.
\end{equation}
\\
\\
The per-crop precision and recall are visualized in \figref{fig:precision} and \figref{fig:recall} (exact numbers are discussed in \apdxref{appendix:labeled_evaluation}). We group the evaluations by the crop season as seasonal variations and challenges can significantly affect model performance. Each test example is put in either the Winter season or Monsoon season depending on the label timestamp. If the label timestamp lies in June-October we put it in the Monsoon group, while labels from November-March are put in the Winter group. Some crops are grown in both seasons, while some are exclusive to one season only. \\
\\
Overall, the performance of the model is significantly better in the winter season compared to monsoon. Within the season, we see that precision improves rapidly for the first 2-3 months as the season progresses, regardless of the season. Some crops, especially rice, are easy to identify early in the season, while others, such as corn and gram, are identified better only later on in the season. Recall also often improves through the season, or stays constant for some crops. We also visualize the overall (macro-average) precision and recall as a function of the minimum prediction confidence through the crop season in \figref{fig:precision_recall_curve}. While the precision at a given confidence level doesn't change much through the season, the recall increases indicating that the model is able to identify more and more crops through the season.\\
\\
Another interesting observation is that the performance at 0 days after the estimated season start is not 0. This is likely because the model is either picking up signals that the season detection missed (meaning that the estimated season start is a little after the actual season start), or the model is using priors from patterns of crop sequences. For instance, it is very common for farmers to grow rice in the monsoon followed by wheat in the winter. The model can exploit such patterns for early season predictions.\\
\\
Notably, the performance of the model is disproportionately poor on the ``others'' category, sorghum, and monsoon corn. This is partly due to a shortage of labels, and also partly because some crop signatures in satellite observations are very similar (\cite{SolerPrezSalazar2021MaizeAS}).

\subsection{Aggregate Evaluation Against Census}
\label{sec:evaluation_with_census}
\begin{figure}[H]
\centering

\begin{subfigure}{.98\textwidth}
  \centering
  \caption{Winter (Rabi)}
  \includegraphics[width=\linewidth]{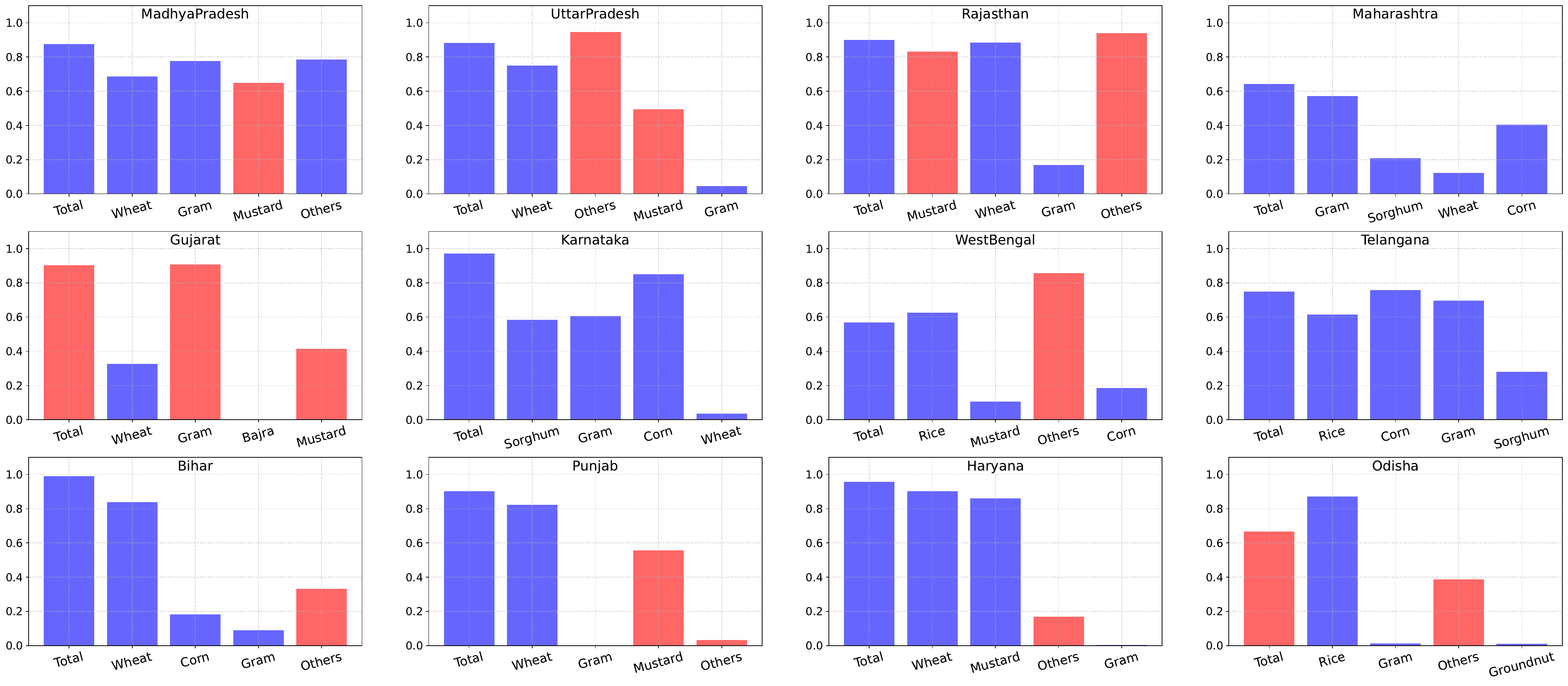}
  \label{fig:normalized_census_winter}
\end{subfigure}%

\begin{subfigure}{.98\textwidth}
  \centering
  \caption{Monsoon (Kharif)}
  \includegraphics[width=\linewidth]{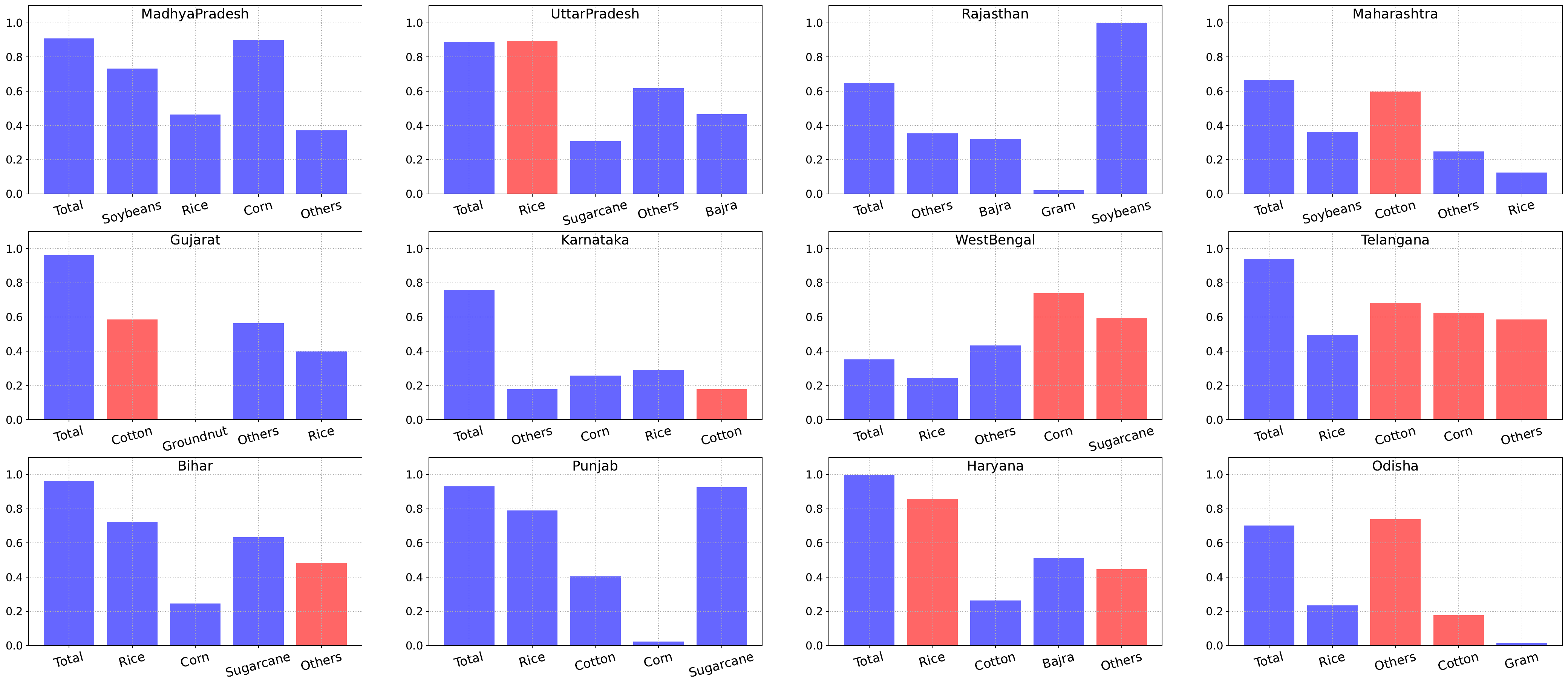}
  \label{fig:normalized_census_monsoon}
\end{subfigure}
\caption[]{\textbf{Evaluation with Census:} \small{Agreement of the predicted area and census area calculated as the minimum of (Predicted Area/ Census Area, Census Area/ Predicted Area) (\eqnref{eqn:census_comparison}). Census 2023-24 used. Twelve largest states by total cultivated area visualized with the major crops per state. Blue bars indicate that the predicted area is less than the census area, and red bars indicate where the predicted area is more than the census area. Note that the total predicted area need not match the census reported area because of either errors in the field boundaries we use, or errors in the census.}}
\label{fig:normalized_census}

\end{figure}


\begin{table}[]
\caption{State-wise Total Census Reported Area, Aggregate Predicted Area, and Cosine Similarity for 2023-24. Areas are in thousand hectares.}
\label{tab:census_area}
\begin{tabular}{@{}lcccccc@{}}
\toprule
 & \multicolumn{3}{c}{\textbf{Winter}} & \multicolumn{3}{c}{\textbf{Monsoon}} \\ \midrule
\multicolumn{1}{|l|}{\textbf{State}} & \multicolumn{1}{c|}{\textbf{Census}} & \multicolumn{1}{c|}{\textbf{Predicted}} & \multicolumn{1}{c|}{\textbf{Cosine Sim}} & \multicolumn{1}{c|}{\textbf{Census}} & \multicolumn{1}{c|}{\textbf{Predicted}} & \multicolumn{1}{c|}{\textbf{Cosine Sim}} \\ \midrule
\multicolumn{1}{|l|}{MadhyaPradesh} & \multicolumn{1}{c|}{13,115} & \multicolumn{1}{c|}{11,468} & \multicolumn{1}{c|}{0.98} & \multicolumn{1}{c|}{14,085} & \multicolumn{1}{c|}{12,811} & \multicolumn{1}{c|}{0.86} \\ \midrule
\multicolumn{1}{|l|}{UttarPradesh} & \multicolumn{1}{c|}{13,230} & \multicolumn{1}{c|}{11,673} & \multicolumn{1}{c|}{0.97} & \multicolumn{1}{c|}{12,343} & \multicolumn{1}{c|}{10,963} & \multicolumn{1}{c|}{0.92} \\ \midrule
\multicolumn{1}{|l|}{Rajasthan} & \multicolumn{1}{c|}{8,678} & \multicolumn{1}{c|}{7,815} & \multicolumn{1}{c|}{0.94} & \multicolumn{1}{c|}{15,882} & \multicolumn{1}{c|}{10,316} & \multicolumn{1}{c|}{0.66} \\ \midrule
\multicolumn{1}{|l|}{Maharashtra} & \multicolumn{1}{c|}{6,285} & \multicolumn{1}{c|}{4,034} & \multicolumn{1}{c|}{0.80} & \multicolumn{1}{c|}{15,607} & \multicolumn{1}{c|}{10,393} & \multicolumn{1}{c|}{0.77} \\ \midrule
\multicolumn{1}{|l|}{Gujarat} & \multicolumn{1}{c|}{2,999} & \multicolumn{1}{c|}{3,324} & \multicolumn{1}{c|}{0.68} & \multicolumn{1}{c|}{7,647} & \multicolumn{1}{c|}{7,359} & \multicolumn{1}{c|}{0.83} \\ \midrule
\multicolumn{1}{|l|}{Karnataka} & \multicolumn{1}{c|}{2,136} & \multicolumn{1}{c|}{2,076} & \multicolumn{1}{c|}{0.78} & \multicolumn{1}{c|}{7,578} & \multicolumn{1}{c|}{5,761} & \multicolumn{1}{c|}{0.38} \\ \midrule
\multicolumn{1}{|l|}{WestBengal} & \multicolumn{1}{c|}{3,208} & \multicolumn{1}{c|}{1,821} & \multicolumn{1}{c|}{0.89} & \multicolumn{1}{c|}{4,957} & \multicolumn{1}{c|}{1,747} & \multicolumn{1}{c|}{0.97} \\ \midrule
\multicolumn{1}{|l|}{Telangana} & \multicolumn{1}{c|}{2,717} & \multicolumn{1}{c|}{2,033} & \multicolumn{1}{c|}{0.98} & \multicolumn{1}{c|}{5,060} & \multicolumn{1}{c|}{4,760} & \multicolumn{1}{c|}{0.87} \\ \midrule
\multicolumn{1}{|l|}{Bihar} & \multicolumn{1}{c|}{3,627} & \multicolumn{1}{c|}{3,590} & \multicolumn{1}{c|}{0.92} & \multicolumn{1}{c|}{3,675} & \multicolumn{1}{c|}{3,545} & \multicolumn{1}{c|}{0.97} \\ \midrule
\multicolumn{1}{|l|}{Punjab} & \multicolumn{1}{c|}{3,634} & \multicolumn{1}{c|}{3,280} & \multicolumn{1}{c|}{1.00} & \multicolumn{1}{c|}{3,596} & \multicolumn{1}{c|}{3,347} & \multicolumn{1}{c|}{0.98} \\ \midrule
\multicolumn{1}{|l|}{Haryana} & \multicolumn{1}{c|}{3,156} & \multicolumn{1}{c|}{3,018} & \multicolumn{1}{c|}{1.00} & \multicolumn{1}{c|}{2,909} & \multicolumn{1}{c|}{2,904} & \multicolumn{1}{c|}{0.94} \\ \midrule
\multicolumn{1}{|l|}{Odisha} & \multicolumn{1}{c|}{915} & \multicolumn{1}{c|}{1,373} & \multicolumn{1}{c|}{0.71} & \multicolumn{1}{c|}{4,730} & \multicolumn{1}{c|}{3,316} & \multicolumn{1}{c|}{0.62} \\ \midrule
\multicolumn{1}{|l|}{Chhattisgarh} & \multicolumn{1}{c|}{596} & \multicolumn{1}{c|}{1,355} & \multicolumn{1}{c|}{0.36} & \multicolumn{1}{c|}{4,220} & \multicolumn{1}{c|}{4,044} & \multicolumn{1}{c|}{0.76} \\ \midrule
\multicolumn{1}{|l|}{TamilNadu} & \multicolumn{1}{c|}{2,583} & \multicolumn{1}{c|}{1,978} & \multicolumn{1}{c|}{0.43} & \multicolumn{1}{c|}{1,871} & \multicolumn{1}{c|}{2,520} & \multicolumn{1}{c|}{0.44} \\ \midrule
\multicolumn{1}{|l|}{AndhraPradesh} & \multicolumn{1}{c|}{1,591} & \multicolumn{1}{c|}{2,634} & \multicolumn{1}{c|}{0.58} & \multicolumn{1}{c|}{2,694} & \multicolumn{1}{c|}{3,630} & \multicolumn{1}{c|}{0.56} \\ \midrule
\multicolumn{1}{|l|}{Assam} & \multicolumn{1}{c|}{934} & \multicolumn{1}{c|}{667} & \multicolumn{1}{c|}{0.40} & \multicolumn{1}{c|}{2,199} & \multicolumn{1}{c|}{931} & \multicolumn{1}{c|}{0.23} \\ \midrule
\multicolumn{1}{|l|}{Jharkhand} & \multicolumn{1}{c|}{1,044} & \multicolumn{1}{c|}{477} & \multicolumn{1}{c|}{0.45} & \multicolumn{1}{c|}{1,764} & \multicolumn{1}{c|}{1,814} & \multicolumn{1}{c|}{0.51} \\ \midrule
\multicolumn{1}{|l|}{Uttarakhand} & \multicolumn{1}{c|}{349} & \multicolumn{1}{c|}{196} & \multicolumn{1}{c|}{0.97} & \multicolumn{1}{c|}{490} & \multicolumn{1}{c|}{248} & \multicolumn{1}{c|}{0.87} \\ \midrule
\multicolumn{1}{|l|}{HimachalPradesh} & \multicolumn{1}{c|}{355} & \multicolumn{1}{c|}{86} & \multicolumn{1}{c|}{0.58} & \multicolumn{1}{c|}{345} & \multicolumn{1}{c|}{123} & \multicolumn{1}{c|}{0.20} \\ \midrule
\multicolumn{1}{|l|}{Nagaland} & \multicolumn{1}{c|}{43} & \multicolumn{1}{c|}{7} & \multicolumn{1}{c|}{0.73} & \multicolumn{1}{c|}{312} & \multicolumn{1}{c|}{12} & \multicolumn{1}{c|}{0.21} \\ \midrule
\multicolumn{1}{|l|}{Tripura} & \multicolumn{1}{c|}{83} & \multicolumn{1}{c|}{30} & \multicolumn{1}{c|}{0.21} & \multicolumn{1}{c|}{210} & \multicolumn{1}{c|}{40} & \multicolumn{1}{c|}{0.25} \\ \midrule
\multicolumn{1}{|l|}{ArunachalPradesh} & \multicolumn{1}{c|}{51} & \multicolumn{1}{c|}{26} & \multicolumn{1}{c|}{0.24} & \multicolumn{1}{c|}{225} & \multicolumn{1}{c|}{38} & \multicolumn{1}{c|}{0.27} \\ \midrule
\multicolumn{1}{|l|}{Manipur} & \multicolumn{1}{c|}{252} & \multicolumn{1}{c|}{13} & \multicolumn{1}{c|}{0.18} & \multicolumn{1}{c|}{17} & \multicolumn{1}{c|}{73} & \multicolumn{1}{c|}{0.37} \\ \midrule
\multicolumn{1}{|l|}{Meghalaya} & \multicolumn{1}{c|}{32} & \multicolumn{1}{c|}{31} & \multicolumn{1}{c|}{0.49} & \multicolumn{1}{c|}{140} & \multicolumn{1}{c|}{45} & \multicolumn{1}{c|}{0.26} \\ \midrule
\multicolumn{1}{|l|}{Sikkim} & \multicolumn{1}{c|}{3} & \multicolumn{1}{c|}{1} & \multicolumn{1}{c|}{0.11} & \multicolumn{1}{c|}{43} & \multicolumn{1}{c|}{9} & \multicolumn{1}{c|}{0.17} \\ \midrule
\multicolumn{1}{|l|}{Mizoram} & \multicolumn{1}{c|}{3} & \multicolumn{1}{c|}{1} & \multicolumn{1}{c|}{0.78} & \multicolumn{1}{c|}{34} & \multicolumn{1}{c|}{10} & \multicolumn{1}{c|}{0.09} \\ \midrule
\textbf{OVERALL} & \textbf{72,160} & \textbf{63,374} & \textbf{0.94} & \textbf{113,328} & \textbf{91,140} & \textbf{0.75} \\ \bottomrule
\end{tabular}
\end{table}
\begin{figure}[H]
\centering

\begin{subfigure}{.98\textwidth}
  \centering
  \caption{Winter (Rabi)}
  \includegraphics[width=\linewidth]{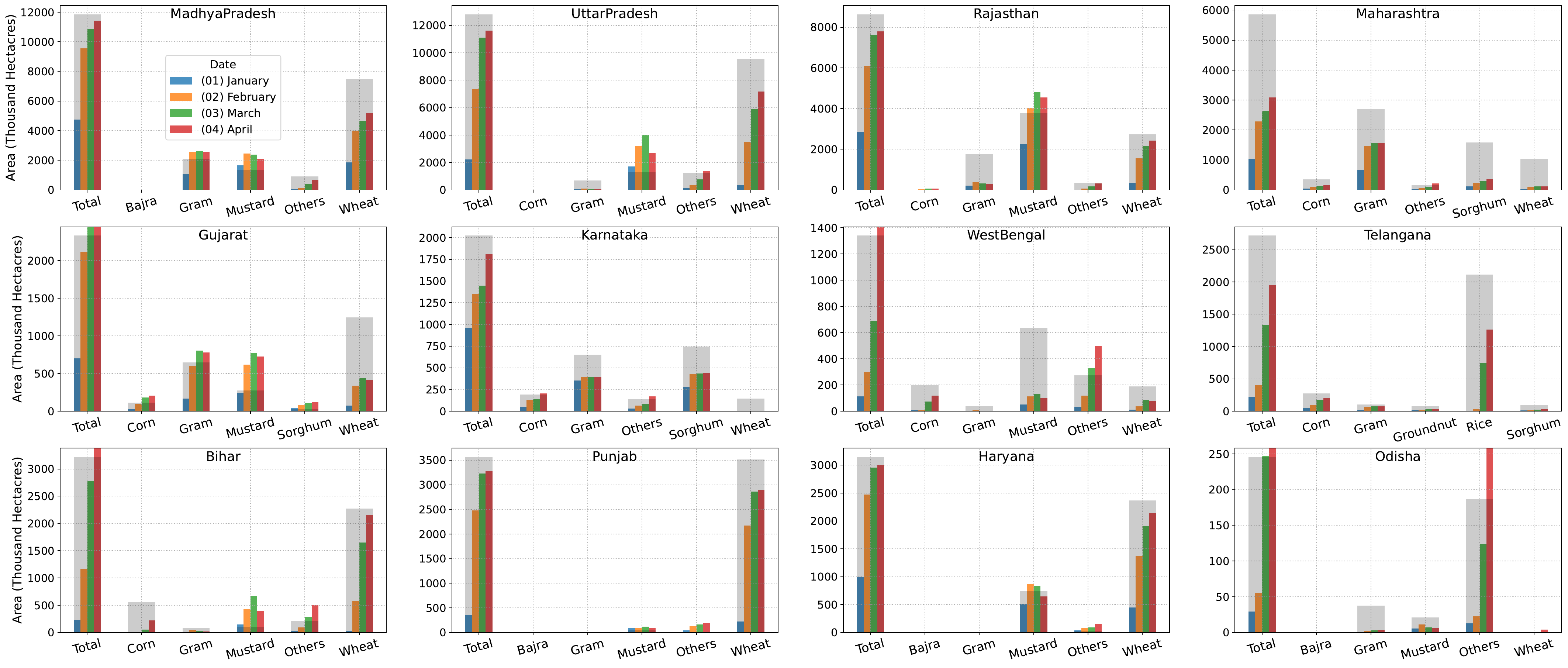}
  \label{fig:census_winter}
\end{subfigure}%

\begin{subfigure}{.98\textwidth}
  \centering
  \caption{Monsoon (Kharif)}
  \includegraphics[width=\linewidth]{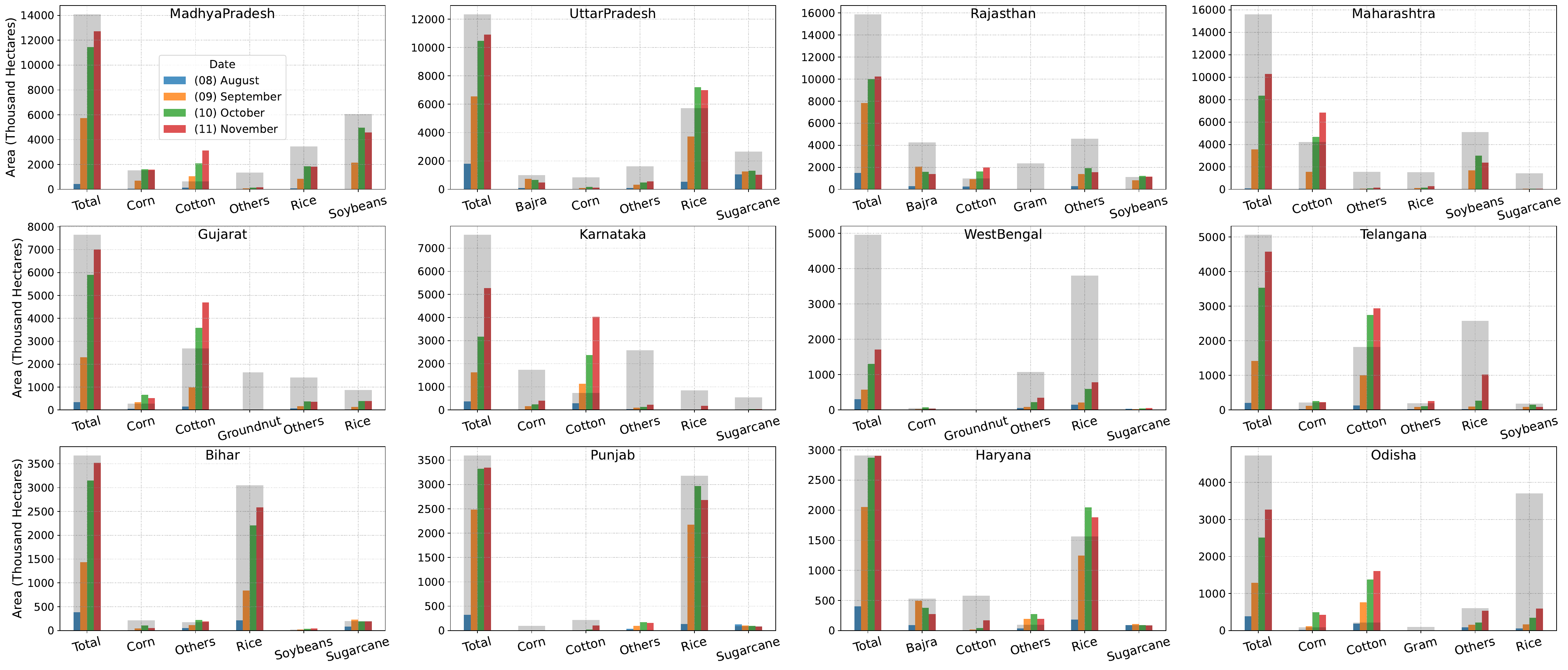}
  \label{fig:census_monsoon}
\end{subfigure}
\caption[]{\textbf{Predicted Area over Time:} \small{The figures visualize the total predicted area under the major crops through the seasons. The gray bars are the Census Reported Area, 2023-24. Twelve largest states by total cultivated area visualized with the major crops per state.}}
\label{fig:census_timeline}

\end{figure}

While the evaluation on the labeled test set does demonstrate the performance of our models, the test set is limited in size. Furthermore it is expensive to increase the size of the labeled test set. Therefore, to scale up our evaluation to cover a larger test set we leverage aggregate labels which are more readily available in the form of census reports. In this section we discuss our evaluation methodology using these aggregate census reports as the ground truth.\\
\\
\noindent The census reports from \cite{UPAg} contain the total area grown under various crops in different administrative regions, grouped by the agricultural season and year. For our analysis we use census reports for the period mid 2023 - mid 2024, aggregated at the administrative region of a state. To evaluate our models against this report, we run inference with our models for all fields and sum up the areas by the predicted crop to generate aggregate predictions. The total area under each crop is calculated as follows:
\begin{equation}
    \text{predicted\_area}[\text{crop}=c] = \sum_{f \in \text{all fields}} \mathbb{1}_{\text{crop}=c}\times \text{area}[f],
\end{equation}
where $\mathbb{1}_{\text{crop}=c}$ is 1 if the most confident prediction for field $f$ is crop $c$, and 0 otherwise.

\noindent Note that since the census reports do not have the specific time for the crop, we cannot perform an exact in-season evaluation here. We therefore evaluate in two ways: (1) we quantitatively compare our latest prediction available for the relevant crop season with the census reports, and (2) we qualitatively show the total identified area per crop over different months. Details of generation of these aggregate predictions can be found in \apdxref{appendix:census_generation}.\\
\\
We perform the quantitative evaluation for 26 states in India (excluded states include Goa for which we do not have ALU fields and Kerala for which we do not have census data available). We compare the areas per crop in the aggregate predictions against the census report for the major crops in each state. The relative distribution of the crops is measured by the cosine similarity~\cite{worrall2023season}. The cosine similarity is defined as:
\begin{equation}
    \text{cosine sim} = \frac{\sum_{c \in \text{crops}} (\text{predicted\_area}[c]*\text{census\_area}[c])}{(\sum_{c \in \text{crops}} \text{predicted\_area}[c])(\sum_{c \in \text{crops}} \text{census\_area}[c])},
\end{equation}
where $\text{predicted\_area}[c]$ is as defined previously, and $\text{census\_area}[c]$ is the total area under crop $c$ reported in the census.\\
\\
\noindent Note that the total agricultural areas in ALU and the census do not match, so the total areas per crop are not expected to match. However, what we want to see here is that the model is able to identify the major crops and the relative distribution, i.e. the cosine similarity, is also high. The comparisons are visualized for the top-12 states (as determined by the total cultivated area reported in the census) in \figref{fig:normalized_census} separately for winter and monsoon. For ease of understanding we visualize the ratio of the predicted and census areas, calculated as:
\begin{equation}
\label{eqn:census_comparison}
    \text{ratio} = \text{min} (\frac{\text{census area}}{\text{predicted area}}, \frac{\text{predicted area}}{\text{census area}}).
\end{equation}
Plots comparing actual areas (in hectares) are available in \figref{fig:census}.
The total reported census area, predicted area, and cosine similarities for all states are documented in \tabref{tab:census_area}.\\
\\
\noindent Similar to the results on labeled data, we see that the aggregate predictions are better for the winter season than for the monsoon. Particularly, the predicted total area for wheat, mustard, and rice match up with the census reports to a reasonable extent during the winter. There are some cotton predictions for winter which is atypical, but those may be due to incorrect season detection or improper season attribution. The results for monsoon are significantly worse. While the model does identify rice, soybeans, and cotton in the appropriate regions, the model over-estimates the area under cotton. This is expected since the model's precision on cotton is relatively poor especially earlier in the season. The discrepancies should resolve with better models and more recent ALU boundaries (since we assume the ALU boundaries are constant indefinitely, which might not be true).\\
\\
\noindent The progression of the total identified area over time is visualized in for the largest 10 agricultural states (again, using the total cultivated area). One observation to note is that the area identified for a given crop doesn't always monotonically increase, which is because the model might update the predictions for a field as it sees more data. Generally for both seasons and most states, we are able to recall around half of the total area that will be eventually identified at least 2 months before the season actually ends.

\section{Future Work}
In the near term, we plan to improve the model performance during the monsoon season. In the long term, we envision building models that jointly solve the field segmentation and crop classification problems. Additionally, we also plan on supporting identification of more crops beyond the 12 we are currently doing.


\newpage
\clearpage
\bibliography{sn-bibliography}

\newpage
\clearpage

\begin{appendices}
\section{Ground Truth Data}
\label{appendix:ground_truth_data}
\subsection{Data Collection Process}

The ground truth labels licensed from partners were collected by following operating procedure:
\begin{itemize}
    \item[1.] Surveyors physically visited fields where there was an active crop.
    \item[2.] The GPS location of the surveyors was recorded from mobile devices they were required to carry.
    \item[3.] Surveyors recorded the label of the crop by observing the crop themselves.
    \item[4.] The time of collection of label was automatically registered.
\end{itemize}

\subsection{Crop Label Distribution}

\begin{figure}[h]
    \centering
    \includegraphics[width=0.8\columnwidth]{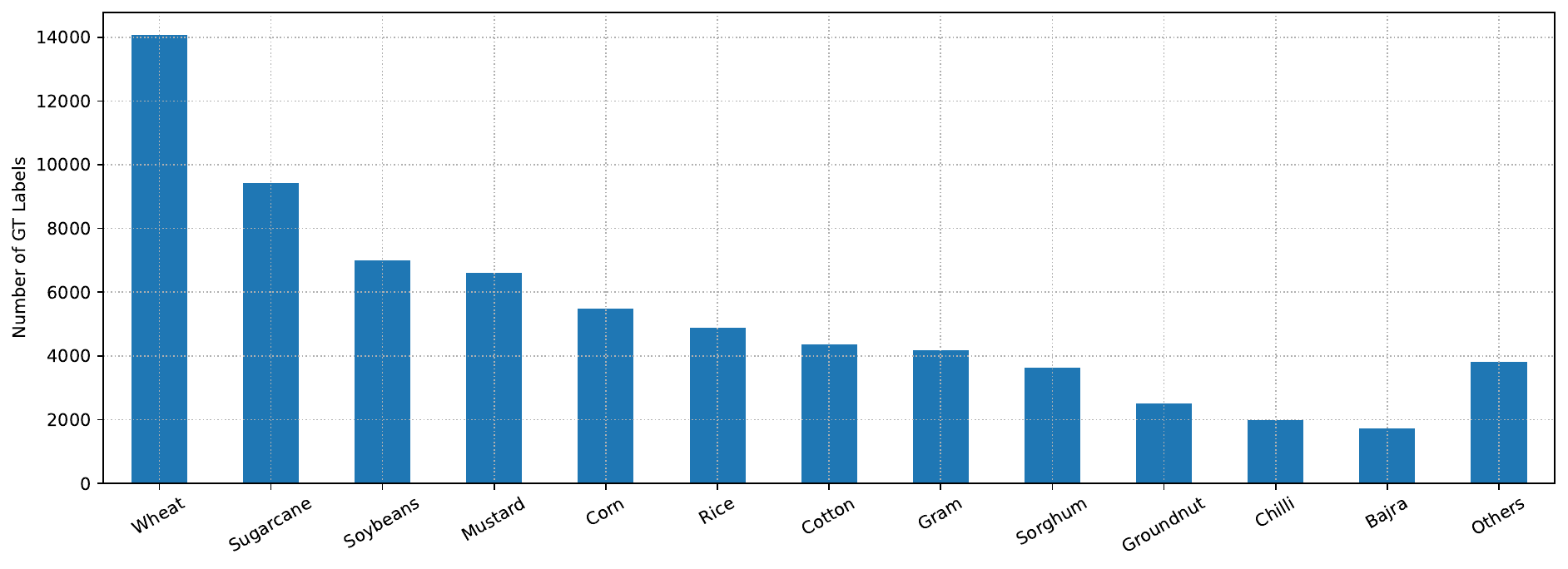}
    \caption{Crop Label Counts}
    \label{fig:crop_label_counts_histogram}
\end{figure}

Distribution of the labels is visualized in \figref{fig:crop_label_counts_histogram}.

\subsection{Geographic Coverage}
\begin{figure}[]
    \centering
    \includegraphics[width=\columnwidth]{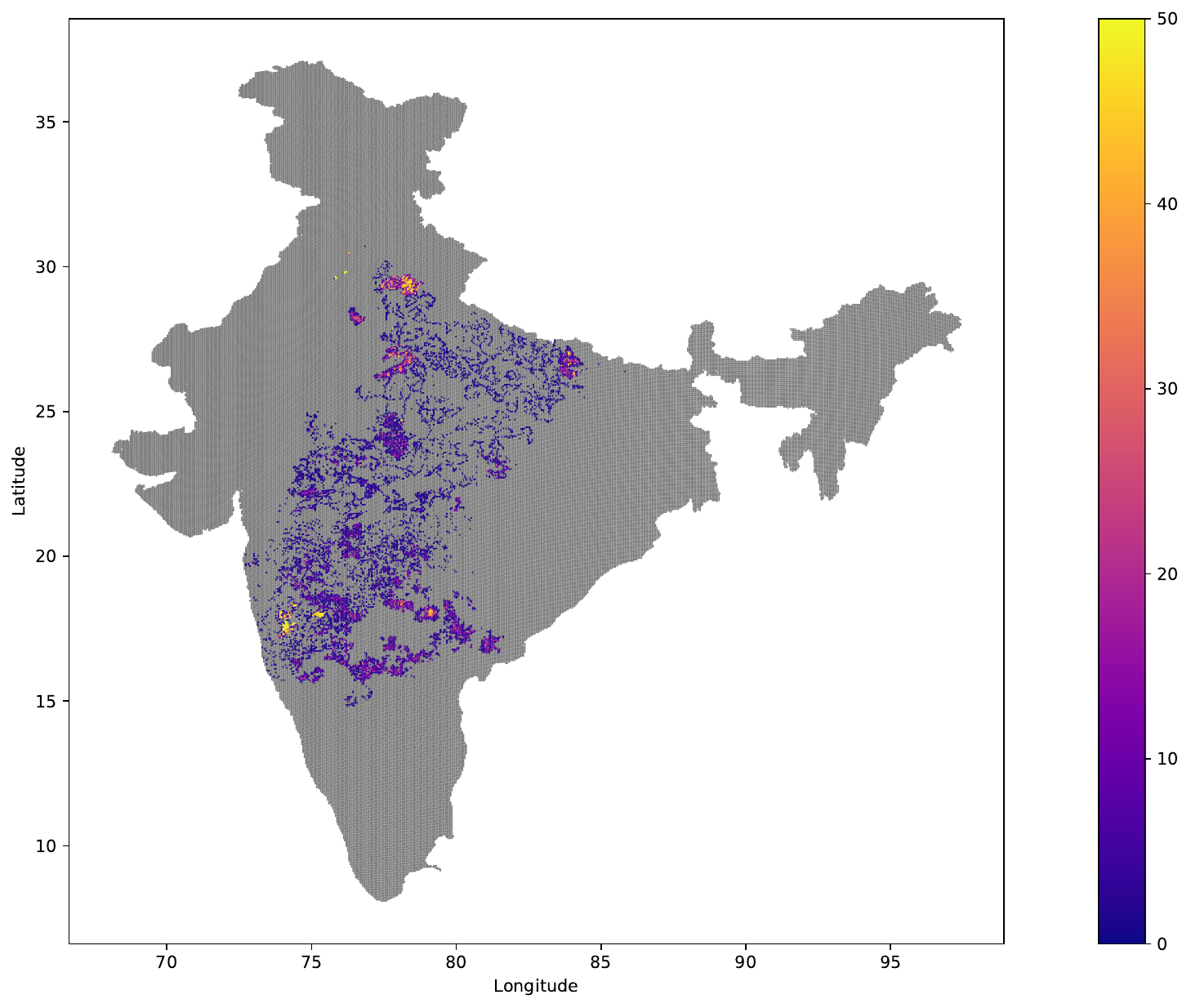}
    \caption{Spatial Distribution of GT Samples Overlayed on Map of India}
    \label{fig:spatial_distribution_of_labels}
\end{figure}

The samples were sourced from 7 states in India, covering $\sim$ 2000 level-9 S2 cells. The spatial distribution of samples is visualized in \figref{fig:spatial_distribution_of_labels}.

\subsection{Temporal Coverage}
\begin{figure}[]
    \centering
    \includegraphics[width=\columnwidth]{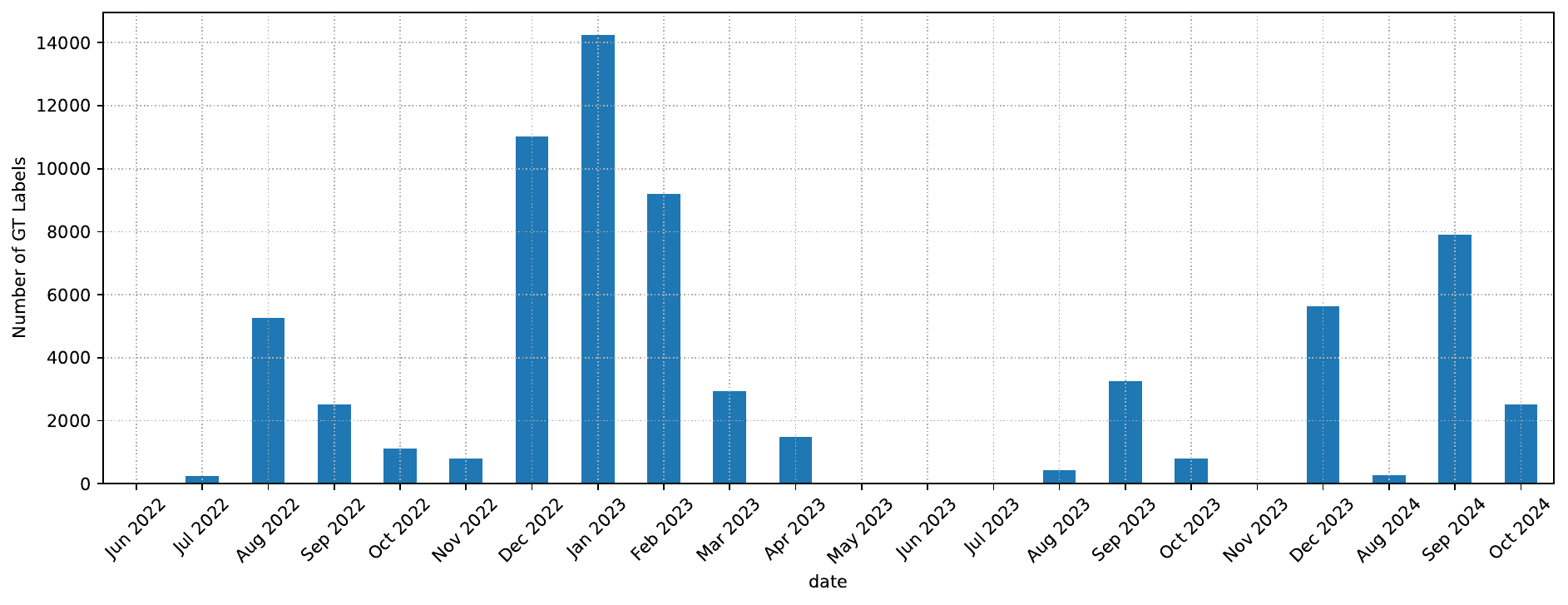}
    \caption{Distribution of Labels Across Time}
    \label{fig:crop_label_temporal_distribution}
\end{figure}

The samples were collected between 2022 - 2024, covering the monsoon and winter seasons. The distribution over time is visualized in \figref{fig:crop_label_temporal_distribution}.

\newpage

\section{Season Detection}
\label{appendix:season_detection}
We use the season detection algorithm from \cite{season_detection} to identify crop seasons and determine the crop season start and end times (roughly the sowing and harvest time). In this section we briefly discuss the algorithm, the inputs used, and the outputs produced. We also present results of estimated crop season lengths for various crops identified by our models.

\subsection{Crop Season Definition}
A crop season is defined here as a time period with the following characteristics:
\begin{itemize}
    \item The start of the season is marked by a change in a field going from bare/ empty state to having vegetation.
    \item The end of the season is marked by the opposite change, i.e. the field going from a vegetated state to a bare/ empty state.
    \item The time between the start and end continuously has vegetation - i.e. there are no further transitions from bare $\Longleftrightarrow$ vegetated between the start and end.
\end{itemize}

Therefore, a crop season in a field is characterized as a tuple of the start and end times $(t_{\text{start}}, t_{\text{end}})$.

\subsection{Algorithm}
The season detection algorithm identifies crop seasons that occur in a field in the following manner:
\begin{itemize}
    \item[1.] We start with Sentinel-2 time-series for the field. We take the median per band across all locations in the field, similar to \secref{sec:training_data_generation}.
    \item[2.] We compute the NDVI using the formula suggested in \cite{kerner2020resilientinseasoncroptype}. This is again a time-series.
    \item[3.] CloudScore+ from Earth Engine (\cite{gorelick2017google}) is used to identify where the NDVI is corrupted by clouds. We drop all NDVI values where the \textbf{cs\_cdf} band from CloudScore+ is $<0.6$, as suggested in the CloudScore+ manual.
    \item[4.] A moving average is applied to the NDVI signal with a window of 20 days.
    \item[5.] The smoothened NDVI is thresholded to get a time-series marking each time step in the time-series as bare or vegetated, where we declare the field to be vegetated at time step $t$ if the corresponding NDVI is $>=0.4$ and bare otherwise. 
    \item[6.] This binary time series is then segmented to find continuous vegetated periods.
    \item[7.] We further apply post-processing to merge two identified seasons if the gap between them is less than 1 month, as this may arise due to signal corruption and is unlikely to happen in the real world.
\end{itemize}

Thus for each field, we get a series of crop seasons, each defined as $(t_{\text{start}}, t_{\text{end}})$.

\subsection{Estimated Crop Season Length}
\begin{table}[]
\caption{Predicted Season Length in Days Per Crop for 2023-24  (shown as percentiles).}
\label{tab:estimated_season_length}
\begin{tabular}{@{}lcccccccc@{}}
\toprule
 & \multicolumn{4}{c}{\textbf{Winter}} & \multicolumn{4}{c}{\textbf{Monsoon}} \\ \midrule
\multicolumn{1}{|l|}{\textbf{Crop}} & \multicolumn{1}{c|}{\textbf{25th }} & \multicolumn{1}{c|}{\textbf{50th }} & \multicolumn{1}{c|}{\textbf{75th }} & \multicolumn{1}{c|}{\textbf{Num Fields}} & \multicolumn{1}{c|}{\textbf{25th }} & \multicolumn{1}{c|}{\textbf{50th }} & \multicolumn{1}{c|}{\textbf{75th }} & \multicolumn{1}{c|}{\textbf{Num Fields}} \\ \midrule
\multicolumn{1}{|l|}{Bajra} & \multicolumn{1}{c|}{90} & \multicolumn{1}{c|}{110} & \multicolumn{1}{c|}{330} & \multicolumn{1}{c|}{21,490} & \multicolumn{1}{c|}{85} & \multicolumn{1}{c|}{95} & \multicolumn{1}{c|}{110} & \multicolumn{1}{c|}{6,390,162} \\ \midrule
\multicolumn{1}{|l|}{Chilli} & \multicolumn{1}{c|}{110} & \multicolumn{1}{c|}{130} & \multicolumn{1}{c|}{145} & \multicolumn{1}{c|}{8,720} & \multicolumn{1}{c|}{160} & \multicolumn{1}{c|}{225} & \multicolumn{1}{c|}{260} & \multicolumn{1}{c|}{4,358} \\ \midrule
\multicolumn{1}{|l|}{Corn} & \multicolumn{1}{c|}{85} & \multicolumn{1}{c|}{105} & \multicolumn{1}{c|}{125} & \multicolumn{1}{c|}{9,591,615} & \multicolumn{1}{c|}{115} & \multicolumn{1}{c|}{130} & \multicolumn{1}{c|}{155} & \multicolumn{1}{c|}{20,824,593} \\ \midrule
\multicolumn{1}{|l|}{Cotton} & \multicolumn{1}{c|}{85} & \multicolumn{1}{c|}{110} & \multicolumn{1}{c|}{150} & \multicolumn{1}{c|}{14,692,940} & \multicolumn{1}{c|}{120} & \multicolumn{1}{c|}{155} & \multicolumn{1}{c|}{210} & \multicolumn{1}{c|}{90,101,714} \\ \midrule
\multicolumn{1}{|l|}{Gram} & \multicolumn{1}{c|}{85} & \multicolumn{1}{c|}{95} & \multicolumn{1}{c|}{110} & \multicolumn{1}{c|}{8,694,711} & \multicolumn{1}{c|}{125} & \multicolumn{1}{c|}{190} & \multicolumn{1}{c|}{220} & \multicolumn{1}{c|}{1,685,755} \\ \midrule
\multicolumn{1}{|l|}{Groundnut} & \multicolumn{1}{c|}{90} & \multicolumn{1}{c|}{105} & \multicolumn{1}{c|}{115} & \multicolumn{1}{c|}{200,168} & \multicolumn{1}{c|}{110} & \multicolumn{1}{c|}{130} & \multicolumn{1}{c|}{205} & \multicolumn{1}{c|}{85,771} \\ \midrule
\multicolumn{1}{|l|}{Mustard} & \multicolumn{1}{c|}{105} & \multicolumn{1}{c|}{120} & \multicolumn{1}{c|}{130} & \multicolumn{1}{c|}{27,572,300} & \multicolumn{1}{c|}{110} & \multicolumn{1}{c|}{135} & \multicolumn{1}{c|}{220} & \multicolumn{1}{c|}{6,724,753} \\ \midrule
\multicolumn{1}{|l|}{Rice} & \multicolumn{1}{c|}{100} & \multicolumn{1}{c|}{110} & \multicolumn{1}{c|}{125} & \multicolumn{1}{c|}{28,939,939} & \multicolumn{1}{c|}{105} & \multicolumn{1}{c|}{120} & \multicolumn{1}{c|}{135} & \multicolumn{1}{c|}{106,252,071} \\ \midrule
\multicolumn{1}{|l|}{Sorghum} & \multicolumn{1}{c|}{80} & \multicolumn{1}{c|}{98} & \multicolumn{1}{c|}{115} & \multicolumn{1}{c|}{2,543,865} & \multicolumn{1}{c|}{115} & \multicolumn{1}{c|}{158} & \multicolumn{1}{c|}{215} & \multicolumn{1}{c|}{822,563} \\ \midrule
\multicolumn{1}{|l|}{Soybeans} & \multicolumn{1}{c|}{80} & \multicolumn{1}{c|}{100} & \multicolumn{1}{c|}{125} & \multicolumn{1}{c|}{216,357} & \multicolumn{1}{c|}{90} & \multicolumn{1}{c|}{95} & \multicolumn{1}{c|}{105} & \multicolumn{1}{c|}{13,672,867} \\ \midrule
\multicolumn{1}{|l|}{Sugarcane} & \multicolumn{1}{c|}{105} & \multicolumn{1}{c|}{135} & \multicolumn{1}{c|}{190} & \multicolumn{1}{c|}{5,118,789} & \multicolumn{1}{c|}{130} & \multicolumn{1}{c|}{190} & \multicolumn{1}{c|}{288} & \multicolumn{1}{c|}{5,926,007} \\ \midrule
\multicolumn{1}{|l|}{Wheat} & \multicolumn{1}{c|}{105} & \multicolumn{1}{c|}{120} & \multicolumn{1}{c|}{135} & \multicolumn{1}{c|}{75,814,355} & \multicolumn{1}{c|}{130} & \multicolumn{1}{c|}{145} & \multicolumn{1}{c|}{215} & \multicolumn{1}{c|}{17,604,151} \\ \bottomrule
\end{tabular}
\end{table}
We show in \tabref{tab:estimated_season_length} the season length for various crops identified by our model. These are based on the predictions made for \secref{sec:evaluation_with_census}.

\section{Crop Season Length}
\label{appendix:season_length}
Each crop has a typical range for the crop season length, i.e. the number of days between sowing and harvesting. We use this knowledge to filter out faulty season detection results while generating training data. Specifically, we define an upper bound on the season length for each crop. When running season detection in training data generation in \secref{sec:generation_process}, if the estimated season length exceeds this value then we do not use that sample. This can be an indication of the season detection algorithm incorrectly combining two crop seasons into a single one. The upper bounds are defined in \tabref{tab:max_season_length}.

\begin{table}[]
\caption{Maximum Allowed Season Length Per Crop}
\label{tab:max_season_length}
\begin{tabular}{@{}lc@{}}
\toprule
\textbf{Crop} & \textbf{Max Season Length (days)} \\ \midrule
\multicolumn{1}{|l|}{Wheat} & \multicolumn{1}{c|}{240} \\ \midrule
\multicolumn{1}{|l|}{Sugarcane} & \multicolumn{1}{c|}{600} \\ \midrule
\multicolumn{1}{|l|}{Soybeans} & \multicolumn{1}{c|}{150} \\ \midrule
\multicolumn{1}{|l|}{Mustard} & \multicolumn{1}{c|}{150} \\ \midrule
\multicolumn{1}{|l|}{Corn} & \multicolumn{1}{c|}{180} \\ \midrule
\multicolumn{1}{|l|}{Rice} & \multicolumn{1}{c|}{200} \\ \midrule
\multicolumn{1}{|l|}{Cotton} & \multicolumn{1}{c|}{300} \\ \midrule
\multicolumn{1}{|l|}{Gram (Bengalgram)} & \multicolumn{1}{c|}{180} \\ \midrule
\multicolumn{1}{|l|}{Sorghum} & \multicolumn{1}{c|}{180} \\ \midrule
\multicolumn{1}{|l|}{Groundnut} & \multicolumn{1}{c|}{180} \\ \midrule
\multicolumn{1}{|l|}{Chilli} & \multicolumn{1}{c|}{240} \\ \midrule
\multicolumn{1}{|l|}{Bajra (Pearl Millet)} & \multicolumn{1}{c|}{150} \\ \bottomrule
\end{tabular}
\end{table}

\section{Model Training}
\label{appendix:model_training}
\subsection{Model Architecture}
\begin{table}[h]
\caption{Hyper-parameters Explored}
\label{tab:hyper-params}
\begin{tabular}{@{}ll@{}}
\toprule
\textbf{Hyper-parameter} & \textbf{Values} \\ \midrule
\multicolumn{1}{|l|}{Token Dimension} & \multicolumn{1}{l|}{16, 64} \\ \midrule
\multicolumn{1}{|l|}{Tokenizer Num Time Steps} & \multicolumn{1}{l|}{4, 8} \\ \midrule
\multicolumn{1}{|l|}{Num Pre-Fusion Encoder Layers} & \multicolumn{1}{l|}{2, 4, 6} \\ \midrule
\multicolumn{1}{|l|}{Num Post-Fusion Encoder Layers} & \multicolumn{1}{l|}{2, 4, 6} \\ \midrule
\multicolumn{1}{|l|}{Num Sentinel-1 Decoder Layers} & \multicolumn{1}{l|}{2, 4, 6} \\ \midrule
\multicolumn{1}{|l|}{Num Sentinel-2 Decoder Layers} & \multicolumn{1}{l|}{2, 4, 6} \\ \midrule
\multicolumn{1}{|l|}{Mask Probability Sentinel-1} & \multicolumn{1}{l|}{0.5, 0.75} \\ \midrule
\multicolumn{1}{|l|}{Mask Probability Sentinel-2} & \multicolumn{1}{l|}{0.5, 0.75} \\ \midrule
\multicolumn{1}{|l|}{Attention Num Heads} & \multicolumn{1}{l|}{4} \\ \midrule
\multicolumn{1}{|l|}{Attention Size} & \multicolumn{1}{l|}{32} \\ \midrule
\multicolumn{1}{|l|}{Classifier MLP Depth} & \multicolumn{1}{l|}{2, 4, 6} \\ \midrule
\multicolumn{1}{|l|}{Classifier MLP Width} & \multicolumn{1}{l|}{64, 512, 1024} \\ \midrule
\multicolumn{1}{|l|}{Classifier Dropout} & \multicolumn{1}{l|}{0, 0.1, 0.2} \\ \midrule
\multicolumn{1}{|l|}{Focal Loss Gamma} & \multicolumn{1}{l|}{0, 1.5, 2.0} \\ \midrule
\multicolumn{1}{|l|}{Adam Learning Rate} & \multicolumn{1}{l|}{0.0002} \\ \bottomrule
\end{tabular}
\end{table}

The model consists of an encoder decoder architecture with an additional head for classification. Details are discussed below. Hyper-parameters explored are listed in \tabref{tab:hyper-params}.\\
\\
\noindent \textbf{Tokenizer:} Each satellite time series is tokenized independently. The tokenizer combines all bands of multiple time-steps together into a single token.\\
\\
\noindent \textbf{Encoder:} The encoder part of the network consists of two separate per-satellite transformer encoders which we refer to as \textit{pre-fusion encoders}, followed by a shared transformer encoder which we refer to as the \textit{post-fusion encoders}. We apply the standard learnable position embedding on the inputs after tokenizing. This is applied once before feeding the pre-fusion encoder, and then once again before feeding the post-fusion encoder. The final output is a latent representation/ embedding. \\
\\
\noindent \textbf{Decoder:} The decoder part consists of independent temporal-broadcast decoders per satellite. The embedding is first ``broadcasted'' to generate a time-series that is aligned with the inputs. Then a learnable position embedding is applied. Finally, a ResNet-like model is applied independently to each time-step.\\
\\
\noindent \textbf{Classification Head:} We use a simple MLP on top of the embeddings for classification with 13 outputs (for the 12 crops + other class).\\

\subsection{Pre-training}
\textbf{Data:} We generate training data for pre-training following Steps 1, 2, and 6 from \ref{sec:training_data_generation}. Since this does not require labels, this data is available from all around India.\\
\\
\noindent \textbf{Method:} The pre-training of the encoder is carried out using the MAE (\cite{he2021maskedautoencodersscalablevision}) framework. We randomly mask a large fraction of the tokens for each satellite independently. The encoder-decoder then reconstruct the entire signal (masked and visible) for all satellites. The model is then trained using the MSE loss for the masked inputs. We use the Adam optimizer (\cite{kingma2017adammethodstochasticoptimization}) for training.\\
\\
\noindent \textbf{Model Selection:} For hyper-parameter and checkpoint selection, we use the training and validation sets of the crop classification dataset. At each validation step we use the encoder to produce latent representations for all examples. We then build a 3-nearest neighbour classifier using the classification training set, and compute the F1-score on the classification validation set. The checkpoint with the highest F1-score is taken.

\subsection{Fine-tuning}
The model is fine-tuned for classification using the focal loss (\cite{lin2018focallossdenseobject}) and trained using the Adam optimizer (\cite{kingma2017adammethodstochasticoptimization}).

\section{Census Generation}
\label{appendix:census_generation}
We run large-scale inference with our models to generate aggregate predictions that we then compare with census reports in \secref{sec:evaluation_with_census}. The exact procedure we use is discussed below:
\begin{itemize}
    \item[1.] We select the administrative boundary of interest (in our case, we use state boundaries). 
    \item[2.] We run our crop classification pipeline (season detection as well as crop classification) for all fields from ALU within this boundary. We use an ensemble of the three models evaluated in the \ref{sec:evaluation}. The prediction probabilities per crop are averaged over the models. 
    \item[3.] The pipeline is run independently for the $1^{st}$ of each month for the desired time period (Jan 2022 - Dec 2024 in our case). When the pipeline is run for date $t$, the models are given access to satellite signals from before date $t$ only.
    \item[4.] The predictions for each field are then combined based on the identified crop seasons. Each prediction contains the season estimate. We group all overlapping season estimates and take the latest prediction from this group.
    \item[5.] The predictions for each field are marked with the season (Winter or Monsoon) by the estimated start date of the season (May-October is Monsoon, November-March is Winter).
    \item[6.] The total area under each crop is then summed up over all fields, grouped by the predicted crop.
\end{itemize}

\begin{figure}[H]
\centering

\begin{subfigure}{.98\textwidth}
  \centering
  \caption{Winter (Rabi)}
  \includegraphics[width=\linewidth]{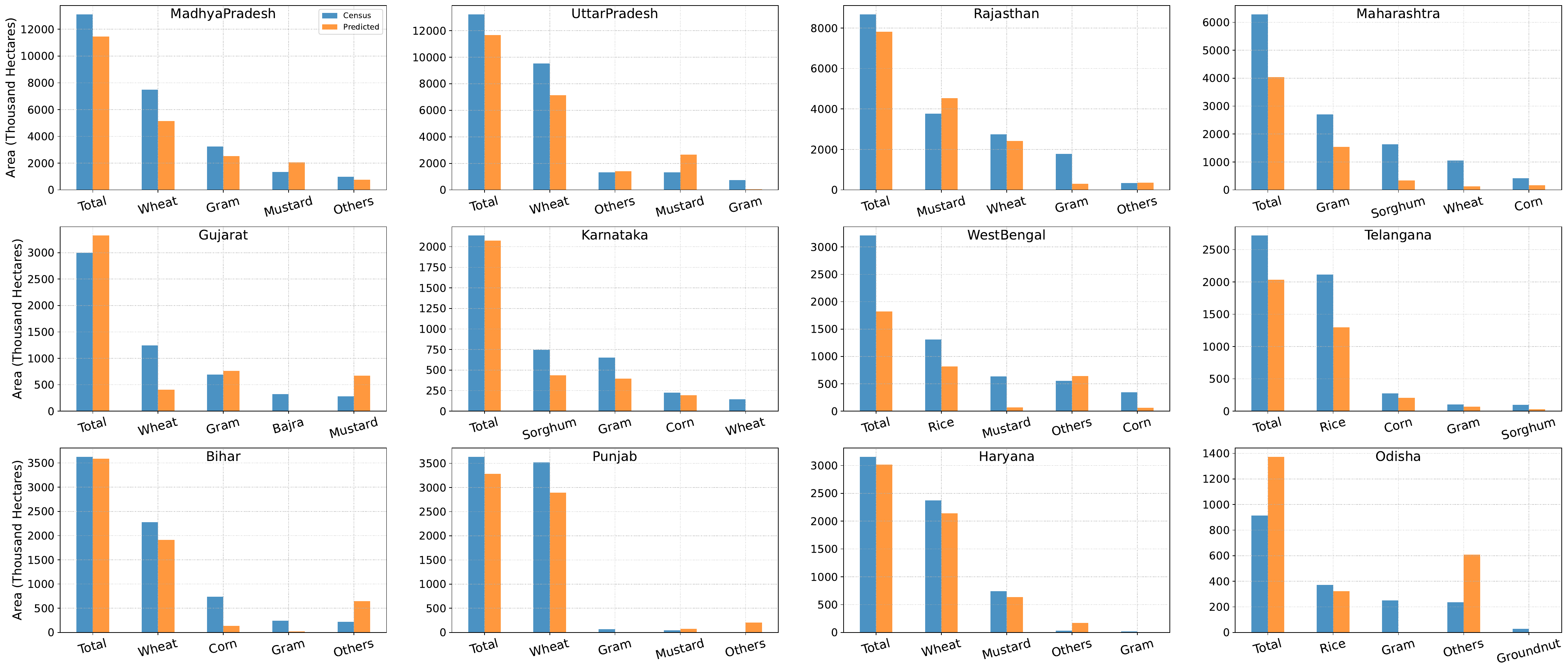}
  \label{fig:census_winter}
\end{subfigure}%

\begin{subfigure}{.98\textwidth}
  \centering
  \caption{Monsoon (Kharif)}
  \includegraphics[width=\linewidth]{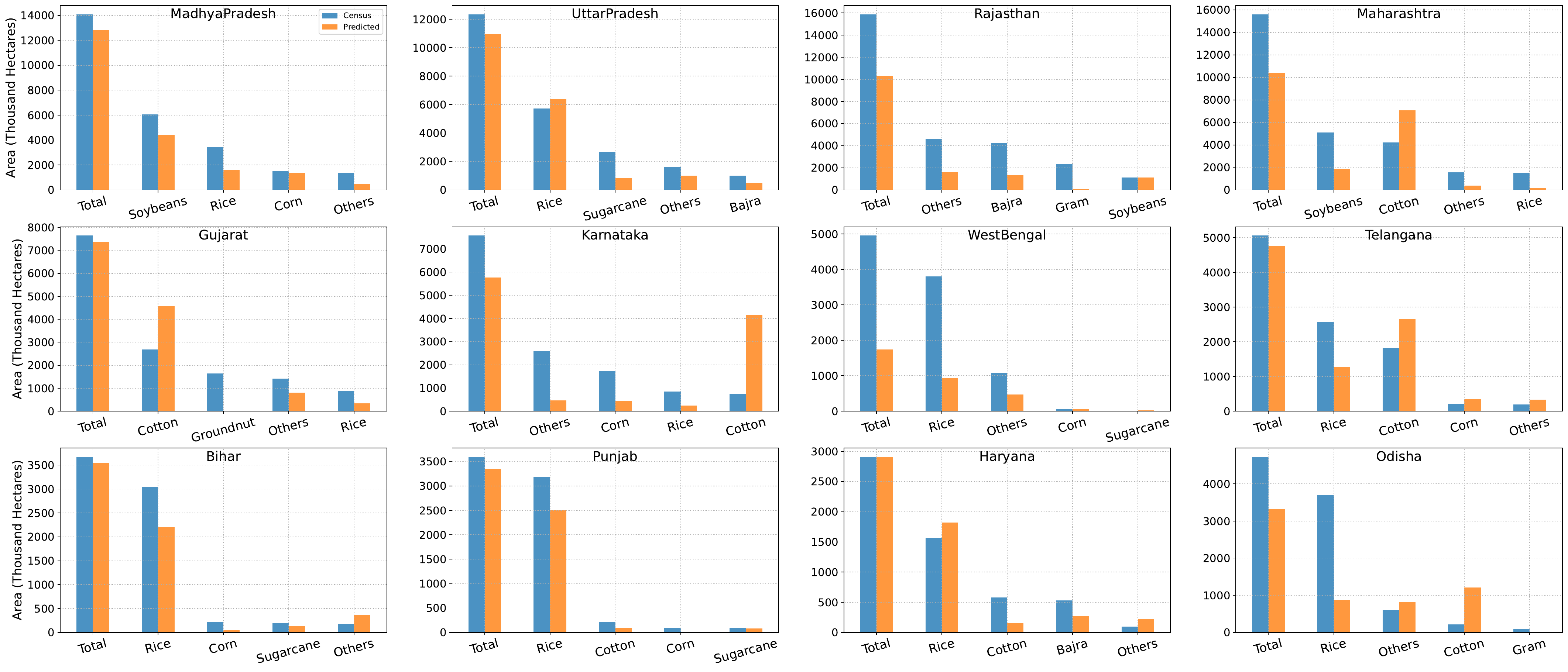}
  \label{fig:census_monsoon}
\end{subfigure}
\caption[]{\textbf{Evaluation with Census:} \small{Final Predicted Area in thousand hectares vs Census Reported Area, 2023-24. Twelve largest states by total cultivated area visualized with the major crops per state. Note that the total predicted area need not match the census reported area because of either errors in the field boundaries we use, or errors in the census.}}
\label{fig:census}

\end{figure}

\section{Evaluation Results}
\label{appendix:labeled_evaluation}
This section contains additional results for \secref{sec:evaluation_with_labels}. We present the numeric values of the metrics, computed as the mean and standard deviation over 3 random splits. The precision per predicted crop in shown in \tabref{tab:precision_winter} and \tabref{tab:precision_monsoon}. The recall in the first prediction is shown  \tabref{tab:recall_winter}, and \tabref{tab:recall_monsoon}. Recall in the top-2 predictions is shown in \tabref{tab:recall_top_2_winter} and \tabref{tab:recall_top_2_monsoon}. ``NaN'' values for precision indicate that no predictions were made. ``NaN'' values for recall indicate that there are no ground truth samples.

\begin{table}[h]
\caption{Precision (Mean, Std) vs Days After Estimated Season Start for Winter}
\label{tab:precision_winter}
\begin{tabular}{@{}llllllll@{}}
\toprule
\textbf{Crop} & \textbf{0 Days} & \textbf{30 Days} & \textbf{60 Days} & \textbf{90 Days} & \textbf{120 Days} & \textbf{150 Days} & \textbf{180 Days} \\ \midrule
\multicolumn{1}{|l|}{Bajra} & \multicolumn{1}{l|}{0.00±0.00} & \multicolumn{1}{l|}{0.00±0.00} & \multicolumn{1}{l|}{0.00±0.00} & \multicolumn{1}{l|}{0.00±0.00} & \multicolumn{1}{l|}{NaN} & \multicolumn{1}{l|}{NaN} & \multicolumn{1}{l|}{NaN} \\ \midrule
\multicolumn{1}{|l|}{Chilli} & \multicolumn{1}{l|}{0.85±0.05} & \multicolumn{1}{l|}{0.87±0.06} & \multicolumn{1}{l|}{0.88±0.06} & \multicolumn{1}{l|}{0.88±0.06} & \multicolumn{1}{l|}{0.92±0.04} & \multicolumn{1}{l|}{0.94±0.04} & \multicolumn{1}{l|}{0.98±0.02} \\ \midrule
\multicolumn{1}{|l|}{Corn} & \multicolumn{1}{l|}{0.54±0.13} & \multicolumn{1}{l|}{0.64±0.10} & \multicolumn{1}{l|}{0.74±0.02} & \multicolumn{1}{l|}{0.79±0.08} & \multicolumn{1}{l|}{0.85±0.05} & \multicolumn{1}{l|}{0.80±0.11} & \multicolumn{1}{l|}{0.00±0.00} \\ \midrule
\multicolumn{1}{|l|}{Cotton} & \multicolumn{1}{l|}{0.48±0.16} & \multicolumn{1}{l|}{0.62±0.10} & \multicolumn{1}{l|}{0.67±0.11} & \multicolumn{1}{l|}{0.68±0.10} & \multicolumn{1}{l|}{0.73±0.09} & \multicolumn{1}{l|}{0.74±0.07} & \multicolumn{1}{l|}{0.80±0.07} \\ \midrule
\multicolumn{1}{|l|}{Gram} & \multicolumn{1}{l|}{0.63±0.09} & \multicolumn{1}{l|}{0.66±0.07} & \multicolumn{1}{l|}{0.73±0.09} & \multicolumn{1}{l|}{0.81±0.08} & \multicolumn{1}{l|}{0.73±0.07} & \multicolumn{1}{l|}{0.17±0.29} & \multicolumn{1}{l|}{0.00±0.00} \\ \midrule
\multicolumn{1}{|l|}{Groundnut} & \multicolumn{1}{l|}{0.77±0.22} & \multicolumn{1}{l|}{0.77±0.21} & \multicolumn{1}{l|}{0.79±0.21} & \multicolumn{1}{l|}{0.85±0.18} & \multicolumn{1}{l|}{0.81±0.27} & \multicolumn{1}{l|}{0.71±0.34} & \multicolumn{1}{l|}{0.00±0.00} \\ \midrule
\multicolumn{1}{|l|}{Mustard} & \multicolumn{1}{l|}{0.64±0.04} & \multicolumn{1}{l|}{0.65±0.07} & \multicolumn{1}{l|}{0.70±0.03} & \multicolumn{1}{l|}{0.74±0.03} & \multicolumn{1}{l|}{0.79±0.04} & \multicolumn{1}{l|}{0.00±0.00} & \multicolumn{1}{l|}{0.00±0.00} \\ \midrule
\multicolumn{1}{|l|}{Others} & \multicolumn{1}{l|}{0.26±0.12} & \multicolumn{1}{l|}{0.29±0.26} & \multicolumn{1}{l|}{0.32±0.16} & \multicolumn{1}{l|}{0.45±0.10} & \multicolumn{1}{l|}{0.38±0.13} & \multicolumn{1}{l|}{0.42±0.07} & \multicolumn{1}{l|}{0.71±0.09} \\ \midrule
\multicolumn{1}{|l|}{Rice} & \multicolumn{1}{l|}{0.62±0.19} & \multicolumn{1}{l|}{0.77±0.18} & \multicolumn{1}{l|}{0.81±0.17} & \multicolumn{1}{l|}{0.83±0.15} & \multicolumn{1}{l|}{0.85±0.11} & \multicolumn{1}{l|}{0.71±0.33} & \multicolumn{1}{l|}{0.58±0.12} \\ \midrule
\multicolumn{1}{|l|}{Sorghum} & \multicolumn{1}{l|}{0.49±0.03} & \multicolumn{1}{l|}{0.51±0.09} & \multicolumn{1}{l|}{0.56±0.02} & \multicolumn{1}{l|}{0.56±0.08} & \multicolumn{1}{l|}{0.48±0.11} & \multicolumn{1}{l|}{0.42±0.11} & \multicolumn{1}{l|}{0.00±0.00} \\ \midrule
\multicolumn{1}{|l|}{Soybeans} & \multicolumn{1}{l|}{0.00±0.00} & \multicolumn{1}{l|}{0.00±0.00} & \multicolumn{1}{l|}{0.00±0.00} & \multicolumn{1}{l|}{0.00±0.00} & \multicolumn{1}{l|}{0.00±0.00} & \multicolumn{1}{l|}{NaN} & \multicolumn{1}{l|}{NaN} \\ \midrule
\multicolumn{1}{|l|}{Sugarcane} & \multicolumn{1}{l|}{0.30±0.11} & \multicolumn{1}{l|}{0.63±0.20} & \multicolumn{1}{l|}{0.87±0.05} & \multicolumn{1}{l|}{0.92±0.02} & \multicolumn{1}{l|}{0.92±0.06} & \multicolumn{1}{l|}{0.95±0.03} & \multicolumn{1}{l|}{0.95±0.02} \\ \midrule
\multicolumn{1}{|l|}{Wheat} & \multicolumn{1}{l|}{0.70±0.06} & \multicolumn{1}{l|}{0.74±0.05} & \multicolumn{1}{l|}{0.78±0.05} & \multicolumn{1}{l|}{0.79±0.06} & \multicolumn{1}{l|}{0.83±0.08} & \multicolumn{1}{l|}{0.92±0.06} & \multicolumn{1}{l|}{0.28±0.17} \\ \bottomrule
\end{tabular}
\end{table}

\begin{table}[h]
\caption{Precision (Mean, Std) vs Days After Estimated Season Start for Monsoon}
\label{tab:precision_monsoon}
\begin{tabular}{@{}llllllll@{}}
\toprule
\textbf{Crop} & \textbf{0 Days} & \textbf{30 Days} & \textbf{60 Days} & \textbf{90 Days} & \textbf{120 Days} & \textbf{150 Days} & \textbf{180 Days} \\ \midrule
\multicolumn{1}{|l|}{Bajra} & \multicolumn{1}{l|}{0.53±0.34} & \multicolumn{1}{l|}{0.66±0.22} & \multicolumn{1}{l|}{0.62±0.23} & \multicolumn{1}{l|}{0.69±0.24} & \multicolumn{1}{l|}{0.63±0.25} & \multicolumn{1}{l|}{NaN} & \multicolumn{1}{l|}{NaN} \\ \midrule
\multicolumn{1}{|l|}{Chilli} & \multicolumn{1}{l|}{0.00±0.00} & \multicolumn{1}{l|}{0.00±0.00} & \multicolumn{1}{l|}{0.00±0.00} & \multicolumn{1}{l|}{0.00±0.00} & \multicolumn{1}{l|}{0.00±0.00} & \multicolumn{1}{l|}{0.00±0.00} & \multicolumn{1}{l|}{0.00±0.00} \\ \midrule
\multicolumn{1}{|l|}{Corn} & \multicolumn{1}{l|}{0.36±0.06} & \multicolumn{1}{l|}{0.43±0.11} & \multicolumn{1}{l|}{0.51±0.11} & \multicolumn{1}{l|}{0.50±0.10} & \multicolumn{1}{l|}{0.49±0.09} & \multicolumn{1}{l|}{0.55±0.43} & \multicolumn{1}{l|}{0.00±0.00} \\ \midrule
\multicolumn{1}{|l|}{Cotton} & \multicolumn{1}{l|}{0.57±0.17} & \multicolumn{1}{l|}{0.58±0.15} & \multicolumn{1}{l|}{0.64±0.11} & \multicolumn{1}{l|}{0.68±0.13} & \multicolumn{1}{l|}{0.73±0.10} & \multicolumn{1}{l|}{0.90±0.06} & \multicolumn{1}{l|}{0.93±0.08} \\ \midrule
\multicolumn{1}{|l|}{Gram} & \multicolumn{1}{l|}{0.00±0.00} & \multicolumn{1}{l|}{0.00±0.00} & \multicolumn{1}{l|}{0.00±0.00} & \multicolumn{1}{l|}{0.00±0.00} & \multicolumn{1}{l|}{0.00±0.00} & \multicolumn{1}{l|}{0.00±0.00} & \multicolumn{1}{l|}{0.00±0.00} \\ \midrule
\multicolumn{1}{|l|}{Groundnut} & \multicolumn{1}{l|}{0.00±0.00} & \multicolumn{1}{l|}{0.00±0.00} & \multicolumn{1}{l|}{0.00±0.00} & \multicolumn{1}{l|}{0.00±0.00} & \multicolumn{1}{l|}{0.00±0.00} & \multicolumn{1}{l|}{NaN} & \multicolumn{1}{l|}{0.00±0.00} \\ \midrule
\multicolumn{1}{|l|}{Mustard} & \multicolumn{1}{l|}{NaN} & \multicolumn{1}{l|}{NaN} & \multicolumn{1}{l|}{NaN} & \multicolumn{1}{l|}{0.00±0.00} & \multicolumn{1}{l|}{0.00±0.00} & \multicolumn{1}{l|}{NaN} & \multicolumn{1}{l|}{NaN} \\ \midrule
\multicolumn{1}{|l|}{Others} & \multicolumn{1}{l|}{0.33±0.14} & \multicolumn{1}{l|}{0.35±0.15} & \multicolumn{1}{l|}{0.37±0.15} & \multicolumn{1}{l|}{0.37±0.14} & \multicolumn{1}{l|}{0.46±0.15} & \multicolumn{1}{l|}{0.54±0.20} & \multicolumn{1}{l|}{0.68±0.16} \\ \midrule
\multicolumn{1}{|l|}{Rice} & \multicolumn{1}{l|}{0.82±0.06} & \multicolumn{1}{l|}{0.86±0.05} & \multicolumn{1}{l|}{0.88±0.08} & \multicolumn{1}{l|}{0.89±0.01} & \multicolumn{1}{l|}{0.88±0.04} & \multicolumn{1}{l|}{0.84±0.09} & \multicolumn{1}{l|}{0.64±0.32} \\ \midrule
\multicolumn{1}{|l|}{Sorghum} & \multicolumn{1}{l|}{0.00±0.00} & \multicolumn{1}{l|}{0.00±0.00} & \multicolumn{1}{l|}{0.00±0.00} & \multicolumn{1}{l|}{0.00±0.00} & \multicolumn{1}{l|}{0.00±0.00} & \multicolumn{1}{l|}{0.00±0.00} & \multicolumn{1}{l|}{0.00±0.00} \\ \midrule
\multicolumn{1}{|l|}{Soybeans} & \multicolumn{1}{l|}{0.67±0.17} & \multicolumn{1}{l|}{0.77±0.12} & \multicolumn{1}{l|}{0.78±0.11} & \multicolumn{1}{l|}{0.77±0.10} & \multicolumn{1}{l|}{0.68±0.12} & \multicolumn{1}{l|}{0.00±0.00} & \multicolumn{1}{l|}{NaN} \\ \midrule
\multicolumn{1}{|l|}{Sugarcane} & \multicolumn{1}{l|}{0.88±0.04} & \multicolumn{1}{l|}{0.86±0.06} & \multicolumn{1}{l|}{0.86±0.05} & \multicolumn{1}{l|}{0.90±0.03} & \multicolumn{1}{l|}{0.92±0.03} & \multicolumn{1}{l|}{0.94±0.02} & \multicolumn{1}{l|}{0.95±0.02} \\ \midrule
\multicolumn{1}{|l|}{Wheat} & \multicolumn{1}{l|}{0.00±0.00} & \multicolumn{1}{l|}{0.00±0.00} & \multicolumn{1}{l|}{0.00±0.00} & \multicolumn{1}{l|}{0.00±0.00} & \multicolumn{1}{l|}{0.00±0.00} & \multicolumn{1}{l|}{0.00±0.00} & \multicolumn{1}{l|}{0.00±0.00} \\ \bottomrule
\end{tabular}
\end{table}
\begin{table}[h]
\caption{Recall At 1st Prediction (Mean, Std) vs Days After Estimated Season Start for Winter}
\label{tab:recall_winter}
\begin{tabular}{@{}llllllll@{}}
\toprule
\textbf{Crop} & \textbf{0 Days} & \textbf{30 Days} & \textbf{60 Days} & \textbf{90 Days} & \textbf{120 Days} & \textbf{150 Days} & \textbf{180 Days} \\ \midrule
\multicolumn{1}{|l|}{Chilli} & \multicolumn{1}{l|}{0.87±0.05} & \multicolumn{1}{l|}{0.91±0.03} & \multicolumn{1}{l|}{0.90±0.04} & \multicolumn{1}{l|}{0.91±0.02} & \multicolumn{1}{l|}{0.93±0.02} & \multicolumn{1}{l|}{0.96±0.01} & \multicolumn{1}{l|}{0.93±0.05} \\ \midrule
\multicolumn{1}{|l|}{Corn} & \multicolumn{1}{l|}{0.49±0.07} & \multicolumn{1}{l|}{0.56±0.02} & \multicolumn{1}{l|}{0.61±0.08} & \multicolumn{1}{l|}{0.67±0.06} & \multicolumn{1}{l|}{0.74±0.10} & \multicolumn{1}{l|}{0.71±0.10} & \multicolumn{1}{l|}{NaN} \\ \midrule
\multicolumn{1}{|l|}{Cotton} & \multicolumn{1}{l|}{0.94±0.05} & \multicolumn{1}{l|}{0.95±0.02} & \multicolumn{1}{l|}{0.96±0.02} & \multicolumn{1}{l|}{0.92±0.03} & \multicolumn{1}{l|}{0.96±0.01} & \multicolumn{1}{l|}{0.97±0.02} & \multicolumn{1}{l|}{0.96±0.02} \\ \midrule
\multicolumn{1}{|l|}{Gram} & \multicolumn{1}{l|}{0.58±0.07} & \multicolumn{1}{l|}{0.62±0.11} & \multicolumn{1}{l|}{0.72±0.08} & \multicolumn{1}{l|}{0.71±0.08} & \multicolumn{1}{l|}{0.49±0.04} & \multicolumn{1}{l|}{0.07±0.12} & \multicolumn{1}{l|}{NaN} \\ \midrule
\multicolumn{1}{|l|}{Groundnut} & \multicolumn{1}{l|}{0.81±0.03} & \multicolumn{1}{l|}{0.82±0.04} & \multicolumn{1}{l|}{0.84±0.03} & \multicolumn{1}{l|}{0.88±0.01} & \multicolumn{1}{l|}{0.83±0.10} & \multicolumn{1}{l|}{0.54±0.28} & \multicolumn{1}{l|}{NaN} \\ \midrule
\multicolumn{1}{|l|}{Mustard} & \multicolumn{1}{l|}{0.62±0.16} & \multicolumn{1}{l|}{0.70±0.13} & \multicolumn{1}{l|}{0.71±0.09} & \multicolumn{1}{l|}{0.72±0.10} & \multicolumn{1}{l|}{0.79±0.07} & \multicolumn{1}{l|}{NaN} & \multicolumn{1}{l|}{NaN} \\ \midrule
\multicolumn{1}{|l|}{Others} & \multicolumn{1}{l|}{0.11±0.08} & \multicolumn{1}{l|}{0.11±0.11} & \multicolumn{1}{l|}{0.14±0.09} & \multicolumn{1}{l|}{0.23±0.06} & \multicolumn{1}{l|}{0.21±0.04} & \multicolumn{1}{l|}{0.38±0.08} & \multicolumn{1}{l|}{0.71±0.13} \\ \midrule
\multicolumn{1}{|l|}{Rice} & \multicolumn{1}{l|}{0.81±0.12} & \multicolumn{1}{l|}{0.88±0.09} & \multicolumn{1}{l|}{0.87±0.08} & \multicolumn{1}{l|}{0.89±0.06} & \multicolumn{1}{l|}{0.91±0.06} & \multicolumn{1}{l|}{0.95±0.06} & \multicolumn{1}{l|}{0.50±0.00} \\ \midrule
\multicolumn{1}{|l|}{Sorghum} & \multicolumn{1}{l|}{0.48±0.08} & \multicolumn{1}{l|}{0.49±0.08} & \multicolumn{1}{l|}{0.60±0.11} & \multicolumn{1}{l|}{0.68±0.08} & \multicolumn{1}{l|}{0.65±0.08} & \multicolumn{1}{l|}{0.48±0.09} & \multicolumn{1}{l|}{NaN} \\ \midrule
\multicolumn{1}{|l|}{Sugarcane} & \multicolumn{1}{l|}{0.17±0.11} & \multicolumn{1}{l|}{0.32±0.16} & \multicolumn{1}{l|}{0.69±0.08} & \multicolumn{1}{l|}{0.78±0.04} & \multicolumn{1}{l|}{0.84±0.04} & \multicolumn{1}{l|}{0.90±0.04} & \multicolumn{1}{l|}{0.92±0.02} \\ \midrule
\multicolumn{1}{|l|}{Wheat} & \multicolumn{1}{l|}{0.70±0.12} & \multicolumn{1}{l|}{0.74±0.13} & \multicolumn{1}{l|}{0.80±0.07} & \multicolumn{1}{l|}{0.85±0.06} & \multicolumn{1}{l|}{0.88±0.05} & \multicolumn{1}{l|}{0.83±0.17} & \multicolumn{1}{l|}{0.29±0.09} \\ \bottomrule
\end{tabular}
\end{table}

\begin{table}[h]
\caption{Recall At 1st Prediction (Mean, Std) vs Days After Estimated Season Start for Monsoon}
\label{tab:recall_monsoon}
\begin{tabular}{@{}llllllll@{}}
\toprule
\textbf{Crop} & \textbf{0 Days} & \textbf{30 Days} & \textbf{60 Days} & \textbf{90 Days} & \textbf{120 Days} & \textbf{150 Days} & \textbf{180 Days} \\ \midrule
\multicolumn{1}{|l|}{Bajra} & \multicolumn{1}{l|}{0.39±0.21} & \multicolumn{1}{l|}{0.44±0.21} & \multicolumn{1}{l|}{0.43±0.22} & \multicolumn{1}{l|}{0.44±0.19} & \multicolumn{1}{l|}{0.38±0.10} & \multicolumn{1}{l|}{NaN} & \multicolumn{1}{l|}{NaN} \\ \midrule
\multicolumn{1}{|l|}{Corn} & \multicolumn{1}{l|}{0.24±0.09} & \multicolumn{1}{l|}{0.23±0.07} & \multicolumn{1}{l|}{0.26±0.06} & \multicolumn{1}{l|}{0.22±0.05} & \multicolumn{1}{l|}{0.20±0.05} & \multicolumn{1}{l|}{0.14±0.10} & \multicolumn{1}{l|}{NaN} \\ \midrule
\multicolumn{1}{|l|}{Cotton} & \multicolumn{1}{l|}{0.80±0.08} & \multicolumn{1}{l|}{0.81±0.04} & \multicolumn{1}{l|}{0.82±0.03} & \multicolumn{1}{l|}{0.82±0.06} & \multicolumn{1}{l|}{0.89±0.02} & \multicolumn{1}{l|}{0.92±0.01} & \multicolumn{1}{l|}{0.88±0.03} \\ \midrule
\multicolumn{1}{|l|}{Others} & \multicolumn{1}{l|}{0.41±0.10} & \multicolumn{1}{l|}{0.46±0.06} & \multicolumn{1}{l|}{0.47±0.06} & \multicolumn{1}{l|}{0.48±0.06} & \multicolumn{1}{l|}{0.55±0.11} & \multicolumn{1}{l|}{0.59±0.10} & \multicolumn{1}{l|}{0.65±0.12} \\ \midrule
\multicolumn{1}{|l|}{Rice} & \multicolumn{1}{l|}{0.79±0.12} & \multicolumn{1}{l|}{0.83±0.10} & \multicolumn{1}{l|}{0.83±0.09} & \multicolumn{1}{l|}{0.81±0.09} & \multicolumn{1}{l|}{0.84±0.06} & \multicolumn{1}{l|}{0.70±0.23} & \multicolumn{1}{l|}{0.44±0.23} \\ \midrule
\multicolumn{1}{|l|}{Soybeans} & \multicolumn{1}{l|}{0.78±0.10} & \multicolumn{1}{l|}{0.75±0.11} & \multicolumn{1}{l|}{0.80±0.07} & \multicolumn{1}{l|}{0.86±0.07} & \multicolumn{1}{l|}{0.74±0.11} & \multicolumn{1}{l|}{NaN} & \multicolumn{1}{l|}{NaN} \\ \midrule
\multicolumn{1}{|l|}{Sugarcane} & \multicolumn{1}{l|}{0.75±0.07} & \multicolumn{1}{l|}{0.81±0.07} & \multicolumn{1}{l|}{0.82±0.08} & \multicolumn{1}{l|}{0.83±0.09} & \multicolumn{1}{l|}{0.87±0.09} & \multicolumn{1}{l|}{0.90±0.07} & \multicolumn{1}{l|}{0.95±0.03} \\ \bottomrule
\end{tabular}
\end{table}

\begin{table}[h]
\caption{Recall In Top-2 Predictions (Mean, Std) vs Days After Estimated Season Start for Winter}
\label{tab:recall_top_2_winter}
\begin{tabular}{@{}llllllll@{}}
\toprule
\textbf{Crop} & \textbf{0 Days} & \textbf{30 Days} & \textbf{60 Days} & \textbf{90 Days} & \textbf{120 Days} & \textbf{150 Days} & \textbf{180 Days} \\ \midrule
\multicolumn{1}{|l|}{Chilli} & \multicolumn{1}{l|}{0.93±0.05} & \multicolumn{1}{l|}{0.96±0.03} & \multicolumn{1}{l|}{0.96±0.02} & \multicolumn{1}{l|}{0.96±0.02} & \multicolumn{1}{l|}{0.97±0.02} & \multicolumn{1}{l|}{0.98±0.01} & \multicolumn{1}{l|}{0.98±0.02} \\ \midrule
\multicolumn{1}{|l|}{Corn} & \multicolumn{1}{l|}{0.73±0.05} & \multicolumn{1}{l|}{0.70±0.08} & \multicolumn{1}{l|}{0.74±0.07} & \multicolumn{1}{l|}{0.82±0.08} & \multicolumn{1}{l|}{0.89±0.05} & \multicolumn{1}{l|}{0.88±0.08} & \multicolumn{1}{l|}{NaN} \\ \midrule
\multicolumn{1}{|l|}{Cotton} & \multicolumn{1}{l|}{0.98±0.02} & \multicolumn{1}{l|}{0.98±0.03} & \multicolumn{1}{l|}{0.98±0.03} & \multicolumn{1}{l|}{0.98±0.03} & \multicolumn{1}{l|}{0.98±0.01} & \multicolumn{1}{l|}{0.98±0.02} & \multicolumn{1}{l|}{0.98±0.00} \\ \midrule
\multicolumn{1}{|l|}{Gram} & \multicolumn{1}{l|}{0.79±0.02} & \multicolumn{1}{l|}{0.84±0.03} & \multicolumn{1}{l|}{0.87±0.06} & \multicolumn{1}{l|}{0.85±0.06} & \multicolumn{1}{l|}{0.74±0.03} & \multicolumn{1}{l|}{0.28±0.30} & \multicolumn{1}{l|}{NaN} \\ \midrule
\multicolumn{1}{|l|}{Groundnut} & \multicolumn{1}{l|}{0.90±0.02} & \multicolumn{1}{l|}{0.89±0.02} & \multicolumn{1}{l|}{0.91±0.03} & \multicolumn{1}{l|}{0.92±0.02} & \multicolumn{1}{l|}{0.91±0.05} & \multicolumn{1}{l|}{0.69±0.17} & \multicolumn{1}{l|}{NaN} \\ \midrule
\multicolumn{1}{|l|}{Mustard} & \multicolumn{1}{l|}{0.92±0.03} & \multicolumn{1}{l|}{0.95±0.01} & \multicolumn{1}{l|}{0.96±0.02} & \multicolumn{1}{l|}{0.96±0.03} & \multicolumn{1}{l|}{0.97±0.02} & \multicolumn{1}{l|}{NaN} & \multicolumn{1}{l|}{NaN} \\ \midrule
\multicolumn{1}{|l|}{Others} & \multicolumn{1}{l|}{0.26±0.08} & \multicolumn{1}{l|}{0.37±0.09} & \multicolumn{1}{l|}{0.48±0.09} & \multicolumn{1}{l|}{0.54±0.02} & \multicolumn{1}{l|}{0.49±0.05} & \multicolumn{1}{l|}{0.77±0.10} & \multicolumn{1}{l|}{0.94±0.08} \\ \midrule
\multicolumn{1}{|l|}{Rice} & \multicolumn{1}{l|}{0.88±0.09} & \multicolumn{1}{l|}{0.89±0.11} & \multicolumn{1}{l|}{0.90±0.10} & \multicolumn{1}{l|}{0.91±0.07} & \multicolumn{1}{l|}{0.92±0.07} & \multicolumn{1}{l|}{0.97±0.03} & \multicolumn{1}{l|}{0.75±0.35} \\ \midrule
\multicolumn{1}{|l|}{Sorghum} & \multicolumn{1}{l|}{0.74±0.07} & \multicolumn{1}{l|}{0.78±0.09} & \multicolumn{1}{l|}{0.84±0.09} & \multicolumn{1}{l|}{0.88±0.03} & \multicolumn{1}{l|}{0.81±0.08} & \multicolumn{1}{l|}{0.70±0.04} & \multicolumn{1}{l|}{NaN} \\ \midrule
\multicolumn{1}{|l|}{Sugarcane} & \multicolumn{1}{l|}{0.41±0.15} & \multicolumn{1}{l|}{0.46±0.13} & \multicolumn{1}{l|}{0.77±0.09} & \multicolumn{1}{l|}{0.84±0.03} & \multicolumn{1}{l|}{0.91±0.02} & \multicolumn{1}{l|}{0.94±0.03} & \multicolumn{1}{l|}{0.98±0.01} \\ \midrule
\multicolumn{1}{|l|}{Wheat} & \multicolumn{1}{l|}{0.92±0.03} & \multicolumn{1}{l|}{0.94±0.04} & \multicolumn{1}{l|}{0.94±0.04} & \multicolumn{1}{l|}{0.95±0.04} & \multicolumn{1}{l|}{0.96±0.02} & \multicolumn{1}{l|}{0.93±0.07} & \multicolumn{1}{l|}{0.69±0.03} \\ \bottomrule
\end{tabular}
\end{table}

\begin{table}[h]
\caption{Recall In Top-2 Predictions (Mean, Std) vs Days After Estimated Season Start for Monsoon}
\label{tab:recall_top_2_monsoon}
\begin{tabular}{@{}llllllll@{}}
\toprule
\textbf{Crop} & \textbf{0 Days} & \textbf{30 Days} & \textbf{60 Days} & \textbf{90 Days} & \textbf{120 Days} & \textbf{150 Days} & \textbf{180 Days} \\ \midrule
\multicolumn{1}{|l|}{Bajra} & \multicolumn{1}{l|}{0.57±0.19} & \multicolumn{1}{l|}{0.59±0.18} & \multicolumn{1}{l|}{0.60±0.17} & \multicolumn{1}{l|}{0.60±0.16} & \multicolumn{1}{l|}{0.52±0.12} & \multicolumn{1}{l|}{NaN} & \multicolumn{1}{l|}{NaN} \\ \midrule
\multicolumn{1}{|l|}{Corn} & \multicolumn{1}{l|}{0.51±0.09} & \multicolumn{1}{l|}{0.49±0.06} & \multicolumn{1}{l|}{0.52±0.08} & \multicolumn{1}{l|}{0.50±0.09} & \multicolumn{1}{l|}{0.50±0.04} & \multicolumn{1}{l|}{0.40±0.16} & \multicolumn{1}{l|}{NaN} \\ \midrule
\multicolumn{1}{|l|}{Cotton} & \multicolumn{1}{l|}{0.91±0.04} & \multicolumn{1}{l|}{0.92±0.03} & \multicolumn{1}{l|}{0.93±0.03} & \multicolumn{1}{l|}{0.93±0.04} & \multicolumn{1}{l|}{0.95±0.02} & \multicolumn{1}{l|}{0.97±0.01} & \multicolumn{1}{l|}{0.93±0.02} \\ \midrule
\multicolumn{1}{|l|}{Others} & \multicolumn{1}{l|}{0.81±0.18} & \multicolumn{1}{l|}{0.84±0.14} & \multicolumn{1}{l|}{0.88±0.09} & \multicolumn{1}{l|}{0.87±0.07} & \multicolumn{1}{l|}{0.90±0.05} & \multicolumn{1}{l|}{0.94±0.02} & \multicolumn{1}{l|}{0.98±0.02} \\ \midrule
\multicolumn{1}{|l|}{Rice} & \multicolumn{1}{l|}{0.87±0.09} & \multicolumn{1}{l|}{0.88±0.08} & \multicolumn{1}{l|}{0.89±0.07} & \multicolumn{1}{l|}{0.87±0.07} & \multicolumn{1}{l|}{0.89±0.06} & \multicolumn{1}{l|}{0.80±0.15} & \multicolumn{1}{l|}{0.54±0.40} \\ \midrule
\multicolumn{1}{|l|}{Soybeans} & \multicolumn{1}{l|}{0.91±0.05} & \multicolumn{1}{l|}{0.90±0.05} & \multicolumn{1}{l|}{0.93±0.05} & \multicolumn{1}{l|}{0.94±0.03} & \multicolumn{1}{l|}{0.88±0.08} & \multicolumn{1}{l|}{NaN} & \multicolumn{1}{l|}{NaN} \\ \midrule
\multicolumn{1}{|l|}{Sugarcane} & \multicolumn{1}{l|}{0.86±0.02} & \multicolumn{1}{l|}{0.91±0.02} & \multicolumn{1}{l|}{0.92±0.02} & \multicolumn{1}{l|}{0.94±0.03} & \multicolumn{1}{l|}{0.95±0.03} & \multicolumn{1}{l|}{0.98±0.01} & \multicolumn{1}{l|}{0.99±0.01} \\ \bottomrule
\end{tabular}
\end{table}


\end{appendices}


\end{document}